\documentclass{article}
\PassOptionsToPackage{authoryear,round}{natbib}
\usepackage[preprint]{neurips_2026}
\usepackage[utf8]{inputenc}
\usepackage[T1]{fontenc}
\usepackage{microtype}

\usepackage{amsmath,amsfonts,bm}

\def\eqref#1{equation~\ref{#1}}

\def\1{\bm{1}}

\DeclareMathAlphabet{\mathsfit}{\encodingdefault}{\sfdefault}{m}{sl}
\SetMathAlphabet{\mathsfit}{bold}{\encodingdefault}{\sfdefault}{bx}{n}

\usepackage{algorithm}
\usepackage{algpseudocode}
\usepackage{graphicx}
\usepackage{booktabs}
\usepackage{multirow}
\usepackage{array}
\usepackage{float}
\usepackage{wrapfig}
\usepackage{needspace}
\usepackage{placeins}
\usepackage{url}
\usepackage{enumitem}
\usepackage{tabularx}
\usepackage{xcolor}
\usepackage{colortbl}
\usepackage{hyperref}
\usepackage{etoolbox}
\hypersetup{
  hidelinks,
  pdftitle={WEFT: Scaling Tool-Use Post-Training for General-Purpose Agents}
}

\definecolor{weftgroupgray}{gray}{0.94}
\title{\weft{}: Scaling Tool-Use Post-Training for General-Purpose Agents}
\author{%
  \textbf{Bo Mao\textsuperscript{1}\thanks{Equal contribution.}\quad
    Hang He\textsuperscript{1,3}\footnotemark[1]\quad
    Linting Wang\textsuperscript{2}\footnotemark[1]\quad
    Lizhi Lin\textsuperscript{5}\footnotemark[1]\quad
    Maosen Zhou\textsuperscript{2}\footnotemark[1]} \\[1pt]
  \textbf{Guanming Liu\textsuperscript{2}\quad
    Jinxiu Liu\textsuperscript{5}\quad
    Tianyu Huai\textsuperscript{1,3}\quad Chaoyun Zhang\textsuperscript{5}\quad
    Bingxuan Li\textsuperscript{5}} \\[1pt]
  \textbf{Kepeng Lei\textsuperscript{5}\quad Guanting Dong\textsuperscript{4}\quad
    Zhou Shao\textsuperscript{5}\thanks{Corresponding authors.}\quad
    Rui Zheng\textsuperscript{5}\quad Hang Yan\textsuperscript{5}} \\[1pt]
  \textbf{Jie Zhou\textsuperscript{1}\footnotemark[2]\quad
    Chengcheng Wan\textsuperscript{1,3}\quad Tao Gui\textsuperscript{2,3}\quad
    Liang He\textsuperscript{1}\quad Xipeng Qiu\textsuperscript{2,3}} \\[4pt]
  {\normalfont
    \textsuperscript{1}East China Normal University\quad
    \textsuperscript{2}Fudan University} \\
  {\normalfont
    \textsuperscript{3}Shanghai Innovation Institute\quad
    \textsuperscript{4}Renmin University of China} \\
  {\normalfont
    \textsuperscript{5}Shanghai Qiji Zhifeng Co., Ltd} \\[4pt]
  {\normalfont\small
    \href{mailto:shaozhou@qijizhifeng.com}{\texttt{shaozhou@qijizhifeng.com}}\quad
    \href{mailto:jzhou@cs.ecnu.edu.cn}{\texttt{jzhou@cs.ecnu.edu.cn}}}%
}
\hypersetup{pdfauthor={Bo Mao, Hang He, Linting Wang, Lizhi Lin, Maosen Zhou, Guanming Liu, Jinxiu Liu, Tianyu Huai, Chaoyun Zhang, Bingxuan Li, Kepeng Lei, Guanting Dong, Zhou Shao, Rui Zheng, Hang Yan, Jie Zhou, Chengcheng Wan, Tao Gui, Liang He, Xipeng Qiu}}

\makeatletter
\renewcommand{\@notice}{}
\patchcmd{\@maketitle}
  {\begin{tabular}[t]{c}\bf\rule{\z@}{24\p@}\@author\end{tabular}}
  {\ifx\@author\@empty\else
     \begin{tabular}[t]{c}\bf\rule{\z@}{24\p@}\@author\end{tabular}%
   \fi}
  {}{\PackageError{weft-arxiv}{Cannot suppress the empty author block}{}}
\makeatother

\newcommand{\weft}{\textsc{Weft}}
\newcommand{\parhead}[1]{\par\noindent\textbf{#1}\enspace\ignorespaces}

\begin{document}
\raggedbottom
\maketitle
\begin{abstract}
Recent efforts to scale tool-use post-training have largely centered on the
synthesis of executable environments, which constitute only one component
of a broader agentic interaction system comprising the environment, task,
agent harness, and evaluator. Scaling environments in isolation, however,
does not guarantee commensurate gains in model performance, because
reliable learning signals depend on coherent interactions among all
components of the agentic interaction system. To address this problem, we
introduce \weft{} (Whole-system Evolution For Tool-use Post-training), which
couples scalable agentic interaction system construction, execution-driven
self-evolution, and stable post-training. \weft{} scales agentic interaction
system construction across environment breadth, task complexity, and
interaction diversity. Execution-driven self-evolution iteratively uses execution
traces and state evidence to attribute failures and revise the responsible
components, with fresh rollouts evaluating the changes and providing
evidence for subsequent evolution rounds. For stable post-training at
scale, \weft{} addresses both optimization and execution reliability:
prefix-preserving sampling retains verified progress and atomic-turn credit
assignment localizes learning signals, while MegaMCP maintains isolated,
recoverable state across concurrent rollouts over shared tool services.
Extensive experiments across various models and benchmarks demonstrate the
effectiveness of \weft{} for tool-use post-training. \weft{}-8B and
\weft{}-14B outperform all evaluated matched-size environment-scaling
baselines on BFCL V4, $\tau^2$-Bench, and Claw-Eval. In particular,
\weft{}-14B improves over Agent-World-14B by 6.41, 2.23, and 12.27
percentage points. \weft{}-35B-A3B further extends these
gains to more challenging long-horizon workflow benchmarks,
including Toolathlon-Verified and AutomationBench.
\end{abstract}

\section{Introduction}
\label{sec:introduction}

Large language model (LLM) agents use tools to interact with stateful
environments and accomplish complex, multi-step tasks. Solving these tasks
requires agents to track environment state changes and adapt their actions
to environment feedback
\citep{zhou2023webarenarealisticwebenvironment,
xie2024osworldbenchmarkingmultimodalagents,
trivedi2024appworldcontrollableworldapps,
yao2024taubenchbenchmarktoolagentuser}. Effective post-training of such
agents depends on diverse interaction trajectories and reliable supervision
based on task outcomes. Recent work has explored the programmatic synthesis
of executable environments to expand the range of tools and task scenarios
available for post-training
\citep{song2026envscalerscalingtoolinteractive,
dong2026agentworldscalingrealworld,
wang2026agentworldmodelinfinitysynthetic,
tu2026scaleenvscalingenvironmentsynthesis}.

However, scaling executable environments alone does not ensure reliable
learning signals. Training experience emerges from an agentic interaction
system comprising the environment, task, agent harness, and evaluator.
An executable environment does not guarantee that tasks are feasible or
their outcomes are correctly evaluated. Task failures may arise from faulty
tool implementations, infeasible tasks, or verification errors, rather than
model limitations alone. Attributing these failures to the policy can
penalize valid behavior and introduce misleading learning signals
\citep{cai2025autoforge,xu2025tinyv}.
Effective scaling therefore requires not only broader environment coverage,
but also continual improvement of the complete agentic interaction system.
Execution experience must support both policy learning and the
identification and revision of problematic components.

\begin{figure}[t]
    \centering
    \includegraphics[width=\textwidth]{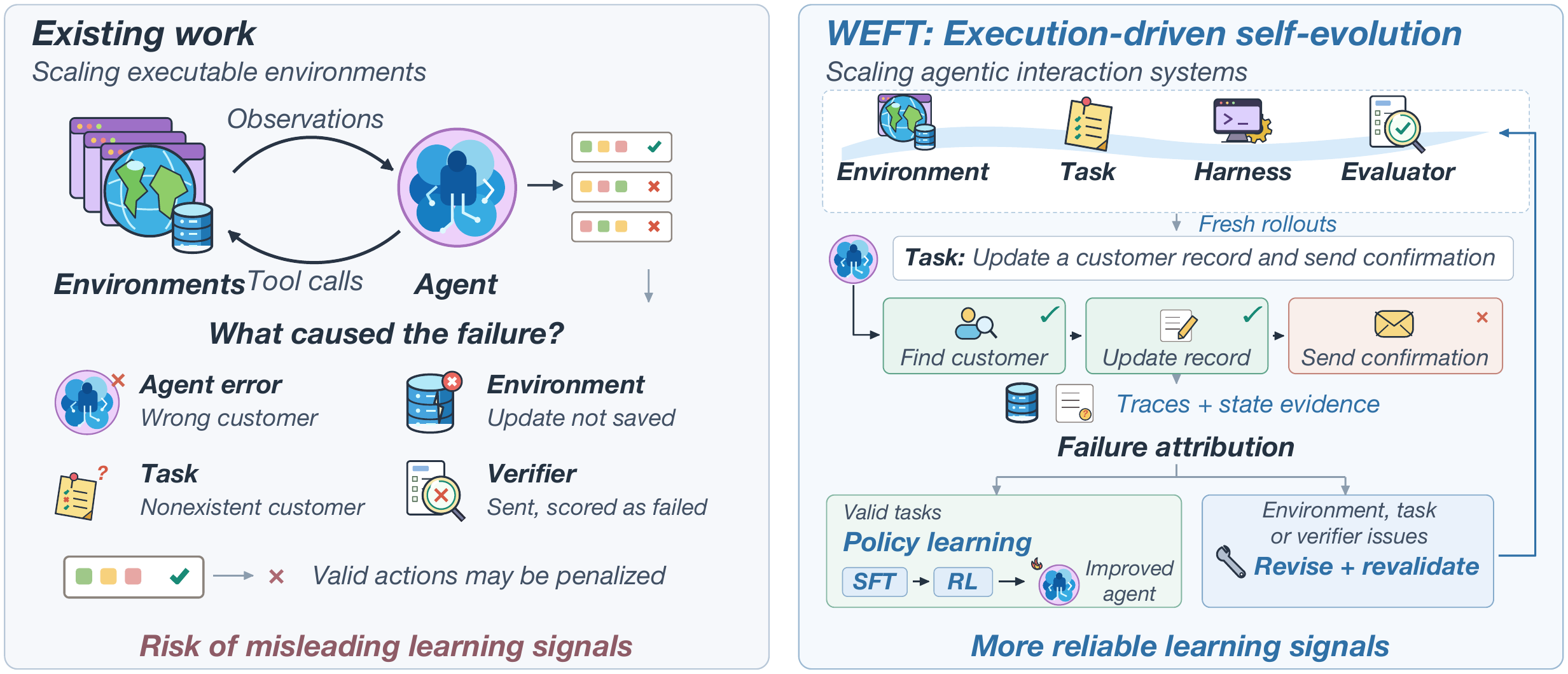}
     \vspace{-4mm}
    \caption{\weft{} improves the agentic interaction system through execution
    feedback. Fresh rollouts evaluate these changes and guide the next round of
    refinement toward more reliable learning signals.}
    \label{fig:motivation}
    \vspace{-3mm}
\end{figure}

To this end, we introduce \weft{}, an end-to-end tool-use post-training
framework integrating scalable agentic interaction system construction,
execution-driven self-evolution, and stable training. It combines stateful
environment synthesis and verifiable cross-tool task composition with
diverse task-disclosure settings and agent harnesses to jointly scale
environment breadth, task complexity, and interaction diversity.
Using execution traces and state evidence, \weft{} distinguishes policy
failures on valid tasks from environment, task, or verifier issues,
guiding targeted component revisions to improve the reliability of
training experience and learning signals at their source. Fresh rollouts
assess revisions and uncover new issues for continued evolution.
For efficient, stable post-training, prefix-preserving rejection sampling
constructs high-quality trajectories by retaining verified progress and
resampling only failed atomic tasks. For reinforcement learning (RL),
task filtering assesses verifier reliability to select training tasks,
while atomic-turn credit assignment provides finer-grained learning signals.
MegaMCP avoids repeated MCP deployment by sharing tool services across
concurrent rollouts. It isolates each rollout's database and workspace state,
using snapshots to support retries and candidate sampling from identical
states.

We post-train Qwen3-8B, Qwen3-14B, and Qwen3.5-35B-A3B with \weft{}
and evaluate the resulting models on five benchmarks spanning tool calling,
multi-turn interaction, and long-horizon task execution. \weft{}-8B and
\weft{}-14B outperform all evaluated matched-size environment-scaling
baselines on the aggregate scores of BFCL V4~\citep{patil2025bfcl},
$\tau^2$-Bench~\citep{barres2025tau2bench}, and Claw-Eval~\citep{ye2026claweval}.
\weft{}-35B-A3B achieves 45.99\% on
Toolathlon-Verified~\citep{li2025toolathlon} and 26.50\% on
AutomationBench~\citep{shepard2026automationbench}. Controlled experiments
further demonstrate the value of continually improving the agentic interaction
system: with the task set and rollout budget fixed, three self-evolution
rounds reduce the tool-call error rate in selected training trajectories by
45.5\% relative to the initial construction and improve three downstream
benchmark scores by 3.65--5.25 percentage points. Ablations also show
performance gains from diverse task-disclosure views and agent harnesses,
prefix-preserving rejection sampling, and atomic-turn credit assignment.
Systems experiments show that MegaMCP reduces data transfer, memory use,
and cold-start latency during concurrent rollouts.

\Needspace{5\baselineskip}
Our main contributions are:
\begin{itemize}[leftmargin=*, align=left, beginpenalty=10000]
    \item We introduce \weft{}, an end-to-end tool-use post-training
    framework that expands the scope of scaling from executable environments
    to complete agentic interaction systems.
    \item We develop execution-driven self-evolution of the
    agentic interaction system for reliable learning signals. Execution
    experience guides failure attribution and targeted component revisions.
    Fresh rollouts assess changes and uncover new issues for subsequent
    evolution rounds.
    \item We support stable post-training through algorithmic
design and execution infrastructure. We develop
efficient prefix-preserving rejection sampling to construct high-quality
trajectories for supervised fine-tuning. In reinforcement learning, task
filtering screens verifier reliability, and atomic-turn credit assignment
provides finer-grained learning signals. For infrastructure, MegaMCP
maintains isolated, recoverable state across concurrent rollouts over
shared tool services.
\end{itemize}

\section{Related Work}
\label{sec:related-work}

\parhead{Environment and task scaling.}
Scaling tool-use training involves both executable environments and the tasks
built within them. Stateful environments support interaction and evaluation
against actual outcomes
\citep{zhou2023webarenarealisticwebenvironment,
xie2024osworldbenchmarkingmultimodalagents,
trivedi2024appworldcontrollableworldapps}.
Automated generation expands environment and compositional task resources
\citep{hu2024agentgen,sullivan2025procedural,shi2025taskcraft}, while
environment-scaling approaches combine programmatic synthesis with task
verification for post-training
\citep{song2026envscalerscalingtoolinteractive,
tu2026scaleenvscalingenvironmentsynthesis,
wang2026agentworldmodelinfinitysynthetic}.
These efforts span workspace tasks \citep{bai2026clawgymscalableframeworkbuilding},
web environments \citep{bai2026webgym,zhang2026infiniteweb}, software engineering
\citep{jain2025r2e,yang2025swe}, and terminal tasks \citep{gandhi2026endless}.
Beyond domain coverage, task construction also varies interaction structure:
real execution traces ground task derivation \citep{chen2026divescalingdiversity},
while topology-aware sampling and structured dependencies support complex
interactions \citep{xu2026envfactory,shi2026scaling}.
\weft{} focuses on cross-MCP task composition, linking execution-verified
atomic tasks through dependencies, entity bindings, and grounded initial
state to support a shared long-horizon goal.

\parhead{Agent interaction and rollout systems.}
Tool-use interaction spans API invocation
\citep{patil2023gorilla,qin2023toolllmfacilitatinglarge}, interleaved reasoning
and action \citep{yao2023reactsynergizingreasoningacting}, and user interaction
in evaluation and data generation
\citep{yao2024taubenchbenchmarktoolagentuser,barres2025tau2bench,
lu2024toolsandbox,prabhakar2025apigenmtagenticpipeline}.
Harness design provides model-oriented computer interfaces
\citep{yang2024swe} and execution and evaluation frameworks
\citep{wang2024openhands,sellier2024browsergymecosystemwebagent}.
Studies compare harness effects under shared tasks
\citep{yao2026harnessbenchmeasuring} and identify performance degradation under
prompt formulations that differ from training \citep{aissi-etal-2025-reinforcement}.
Training systems explore mixed-harness rollouts \citep{song2026clawgym}
and decouple agent execution from optimization
\citep{luo2025agentlightningtrainanyai}.
\weft{} treats interaction diversity and execution support together: the same
grounded task is presented through distinct disclosure formats and native
harnesses, while MegaMCP shares tool services across rollouts with isolated,
recoverable database and workspace state.

\parhead{Self-improvement.}
Self-improvement methods differ in what they revise.
Behavioral approaches refine outputs through self-feedback
\citep{NEURIPS2023_91edff07}, use verbal reflection to guide subsequent attempts
\citep{NEURIPS2023_1b44b878}, or accumulate executable skills through exploration
\citep{wang2023voyageropenendedembodiedagent}.
Task adaptation targets capability gaps \citep{dong2026agentworldscalingrealworld},
adjusts difficulty \citep{zeng2025rlve}, or jointly trains an environment
designer and a solver \citep{liu2026spade}.
System-level approaches search over agent programs
\citep{hu2025automateddesignagenticsystems}, revise harnesses using execution
traces \citep{lin2026agentic}, and coordinate the synthesis, auditing, and repair
of databases, environments, and trajectories \citep{chen2026eigendata}.
\weft{} focuses on attributing execution failures to the components that need
revision, distinguishing policy failures on valid tasks from defects in
environments, tasks, or verifiers.

\parhead{Tool-use post-training.}
Trajectory synthesis combines data generation with quality control.
Rule-based checks, execution, and semantic verification filter training data
\citep{liu2024apigenautomatedpipeline,liu2024toolace}.
Tool relationships and graph structure guide coherent multi-turn dialogue
synthesis \citep{wang2024toolflow,yin2025magnet}, while trajectories containing
feedback and correction improve generalization \citep{fu2025agentrefine}.
For trajectory collection, \weft{} preserves verified prefixes and their
environment state across local failures, applying rejection sampling at
atomic-task boundaries.

Reinforcement learning raises questions of optimization stability, reward
reliability, and credit assignment. Work on group-relative optimization
\citep{shao2024deepseekmathpushinglimitsmathematical} and multi-turn training
stability \citep{wang2025ragenunderstandingselfevolutionllm} is complemented by
studies of reward errors under imperfect verifiers \citep{cai2025reinforcement}.
Credit assignment methods differ in their comparison units: GiGPO groups actions
by recurring states across trajectories \citep{feng2025gigpo}, PivotRL samples
local continuations at informative intermediate turns in demonstrations
\citep{yi2026pivotrl}, and BPO compares sibling terminal returns after branching
from sandbox snapshots \citep{he2026branching}.
\weft{} instead compares candidate atomic turns from identical histories and
environment states, computing group-relative advantages from the current
task's binary completion outcome rather than terminal returns.

\section{Method}
\label{sec:method}

\begin{figure}[t]
    \centering
    \setlength{\parskip}{0pt}
    \setlength{\abovecaptionskip}{4pt}
    \includegraphics[width=\linewidth]{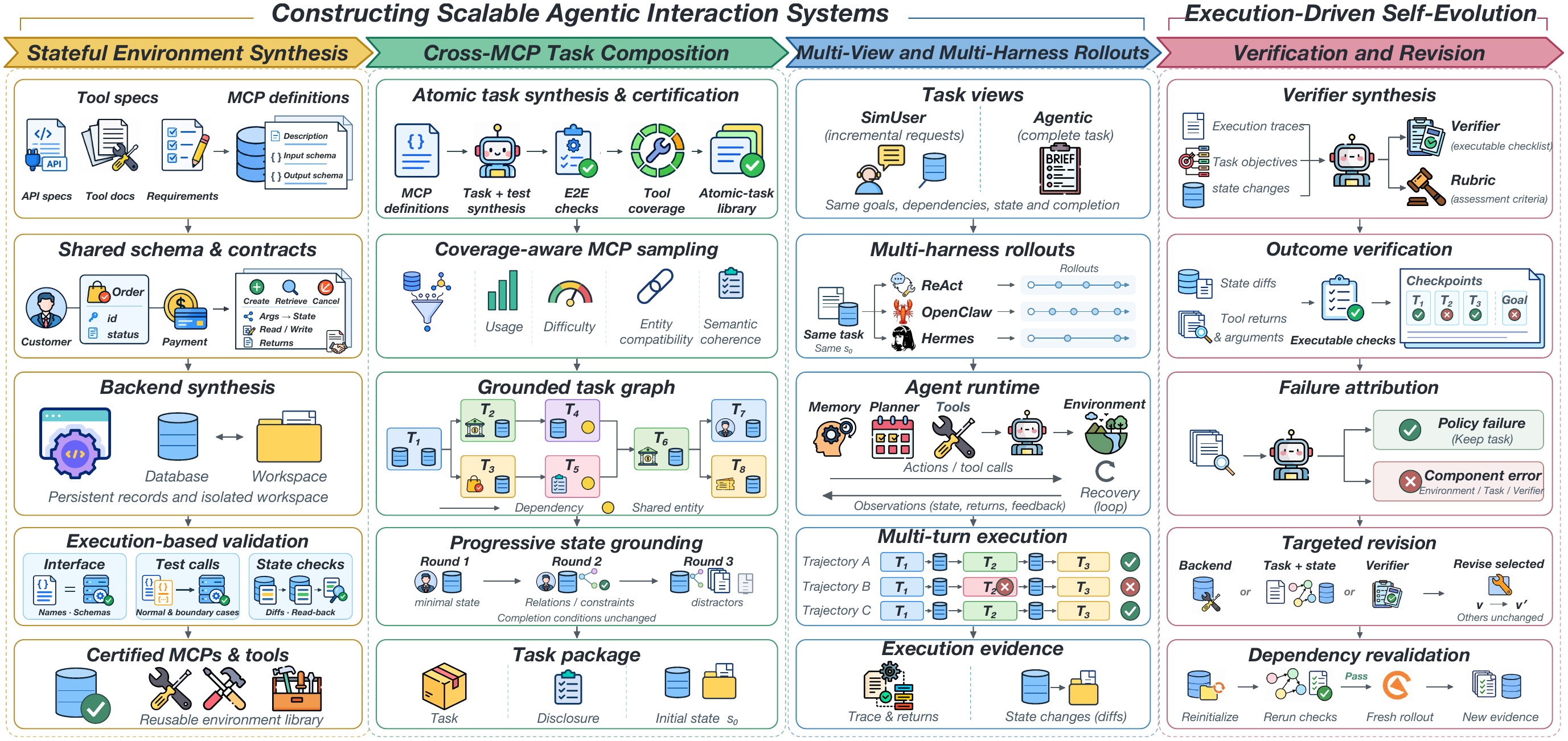}
    \par\nointerlineskip
    \caption{Construction and execution-driven self-evolution in \weft{}.
    Certified atomic tasks form grounded cross-MCP workflows with resettable
    state. Views and harnesses diversify rollouts, whose traces and state
    evidence guide verification, failure attribution, and iterative component
    revision.}
    \label{fig:weft-overview}
    \vspace{-4mm}
\end{figure}

\weft{} organizes tool-use post-training around the agentic interaction
system, with execution supporting both policy learning and continual system
improvement. We jointly expand environment breadth, task complexity, and
interaction diversity through stateful environment synthesis, verifiable
task composition, and rollouts across multiple task views and agent
harnesses (Section~\ref{sec:interaction-framework}). Execution traces and
state evidence support outcome verification, failure attribution, and
targeted component revision; fresh rollouts with the updated system assess
the changes and inform subsequent revisions
(Section~\ref{sec:evaluation}; Figure~\ref{fig:weft-overview}). To support
stable post-training on these interactions, we combine prefix-preserving
rejection sampling, task filtering, and atomic-turn credit assignment with
MegaMCP's isolated, recoverable concurrent execution
(Section~\ref{sec:task-adaptive-training}).

\suppressfloats[t]

\subsection{Constructing Scalable Agentic Interaction Systems}
\label{sec:interaction-framework}

Scaling agentic interaction systems requires not only diverse tool
environments, but also coherent tasks and varied interaction settings.
\weft{} synthesizes stateful environments from tool specifications, generates
and validates reusable atomic tasks, and composes them into cross-MCP
workflows with explicit dependencies and grounded initial states. We vary
task-disclosure settings and agent harnesses while preserving task objectives
and completion conditions. Together, these choices \mbox{expand} environment
breadth, task complexity, and interaction diversity. The resulting rollouts
provide execution traces and state evidence for subsequent verification and
self-evolution.

\subsubsection{Stateful Environment Synthesis}
\label{sec:environment-construction}

\parhead{Shared state and execution contracts.}
We construct MCP tool definitions from public tool documentation and
business requirements, specifying descriptions and input and output schemas.
We define two types of state for tool operations, \emph{database state} and
\emph{workspace state}; tools may operate on either or both. An environment
synthesis agent infers shared entities, read and write
dependencies, and state transitions from the complete tool set. It
instantiates a shared database schema to validate types and relational
constraints. The schema and MCP definitions guide contracts specifying
argument-to-state mappings, reads, writes, and returns; these artifacts
jointly guide FastAPI backend generation. For workspace state, existing
file-oriented runtimes bind to isolated workspaces.

\parhead{Execution-based validation.}
Running interfaces are checked against MCP tool names and argument schemas,
then tested with generated normal and boundary calls. Queries are checked
against state; updates against database or workspace differences and
subsequent reads. Structural, mapping, and runtime or state-access errors
trigger schema, contract, and backend revisions, respectively. Backends
passing revalidation become independently executable MCP servers.

\subsubsection{Cross-MCP Task Composition}
\label{sec:task-world-composition}

\parhead{Certified atomic tasks.}
A task synthesis agent uses each certified MCP's definitions, schema, and
contracts to generate \emph{atomic tasks} and end-to-end (E2E) tests. An
atomic task is an MCP's smallest independently executable and verifiable
unit, realizing one intent through bounded tool calls. Construction uses
\emph{(i)~E2E execution validation}, checking completion against returns and
persistent state after real calls from independent initial states;
\emph{(ii)~tool-chain orthogonality}, pruning identical or contained ordered
chains; \emph{(iii)~tool coverage}, measuring the fraction of MCP tools
covered by retained chains. Failures, redundancy, and uncovered tools guide
incremental generation and revision of the reusable task and test library,
preserving validated tasks (Appendix~\ref{app:weft-atomic-task-certification}).

\parhead{Coverage-aware sampling.}
Configurable profiles specify ranges for atomic-task count, MCP count, and
domain breadth; corpus proportions set generation quotas. Sampling favors
underrepresented domains and MCPs using capacity and usage statistics,
reserving surplus capacity for composition. Candidates are ranked by shared
entities, complementary operations, and functional redundancy. An evaluation
model checks MCP descriptions for a coherent business objective. Usage is
updated before sampling tasks from accepted combinations; rejections trigger
resampling within the profile (Appendix~\ref{app:weft-task-sampling}).

\parhead{Grounded task graph.}
A composition agent links selected atomic tasks into a directed acyclic
graph for a shared business objective: nodes specify intents, inputs, and
outcomes; edges encode information and state dependencies. Entity bindings
align business objects across MCPs.
The graph distinguishes given information from information retrieved through
tools. Deterministic checks ensure nodes use selected tasks, bindings are
conflict-free, and inputs come from initial state or upstream outcomes.
A stable topological order governs disclosure and execution.

\parhead{Progressive state grounding.}
From the graph and MCP schemas, contracts, or workspace structures, a
state-initialization agent builds initial state in three rounds:
\emph{(i)~entity initialization} establishes target entities and minimal
executable state; \emph{(ii)~relation completion} adds cross-tool relations
and constraints; \emph{(iii)~context enrichment} adds historical records and
distractors. All rounds preserve entity correspondence and completion
conditions. States passing schema and task-reachability checks become
resettable initial snapshots.

\subsubsection{Multi-View and Multi-Harness Rollouts}
\label{sec:harnessed-rollout}

\parhead{Task views.}
We vary request and information disclosure while keeping task objectives,
dependencies, initial state, completion conditions, and MCP interfaces fixed
across views and harnesses. Both views retain user-known inputs and upstream
information.
\emph{(i)~Agentic mode} provides a complete brief for autonomous planning.
\emph{(ii)~SimUser mode} uses a user-simulation agent with persona sampling
to disclose requests in dependency order. Each atomic task defines an
\emph{atomic-task turn} allowing multiple tool calls. A final reply without
tool calls ends the turn; SimUser then issues the next request. Independently,
\emph{vague}, \emph{partial}, and \emph{full} vary disclosure explicitness.

\parhead{Harness scaling.}
Our infrastructure routes the same task to ReAct, OpenClaw, or Hermes for
sandboxed rollouts, retaining each harness's prompts, memory, planning loop,
tool-call protocol, and error recovery. ReAct supports both views; OpenClaw
and Hermes use Agentic mode.

\subsection{Execution-Driven Self-Evolution}
\label{sec:evaluation}

Rollouts across task views and agent harnesses produce training trajectories
and can also reveal environment, task, or verifier problems missed by
construction checks. \weft{} uses execution traces and state evidence to
verify task outcomes, distinguish policy errors from component problems,
and revise the affected components. Fresh rollouts with the updated
agentic interaction system evaluate these revisions and inform the next
round, improving the reliability of learning signals through repeated
execution and revision.

\subsubsection{Verification from Execution Evidence}
\label{sec:execution-evidence-verification}

Verification checks atomic-task completion using \emph{(i)~state changes},
captured by database or workspace differences; \emph{(ii)~tool returns}.
Given completion conditions and a reference execution accepted by a
semantic evaluation model, a verifier-generating agent produces and saves
deterministic checks and a corresponding rubric. The checks are tested on
reference evidence and revised using execution feedback; subsequent
rollouts need not follow the reference tool-call sequence.

For trajectory $\tau$, $K$ checkpoints cover atomic tasks and the overall
objective. Each $r_k(\tau)$ is $1$ when all checkpoint conditions hold and
$0$ otherwise; the trajectory score is their mean:
\begin{equation}
    R(\tau)=\frac{1}{K}\sum_{k=1}^{K}r_k(\tau).
    \label{eq:checkpoint-reward}
\end{equation}
\subsubsection{Attribution-Guided Revision}
\label{sec:recursive-refinement}

\parhead{Failure attribution.}
An \emph{attribution agent} examines task graphs, traces, checkpoint outcomes,
and state evidence to distinguish policy failures on valid tasks from
environment, task, or verifier problems. Patterns across views and harnesses
help localize failures; disagreements between rubric-based and executable
scores guide verifier revision. Policy failures alone leave tasks unchanged.

\parhead{Targeted component revision.}
We route environment problems to backends, task or state problems to task
graphs and initial states, and evaluation problems to verifiers. An
\emph{evolution agent} uses the current version and supporting evidence to
revise only the selected components. Task semantics remain fixed across
views within a version.

\parhead{Dependency revalidation and iterative refinement.}
We re-instantiate affected environments and task states and rerun
construction checks. Fresh rollouts on passing versions assess repairs and
reveal unresolved or new problems. Alternating execution and revision drives
iterative evolution of the agentic interaction system through its own
execution experience.

\subsection{Stable Tool-Use Post-Training}
\label{sec:task-adaptive-training}

\begin{figure}[t]
    \centering
    \setlength{\parskip}{0pt}
    \setlength{\abovecaptionskip}{4pt}
    \includegraphics[width=\linewidth]{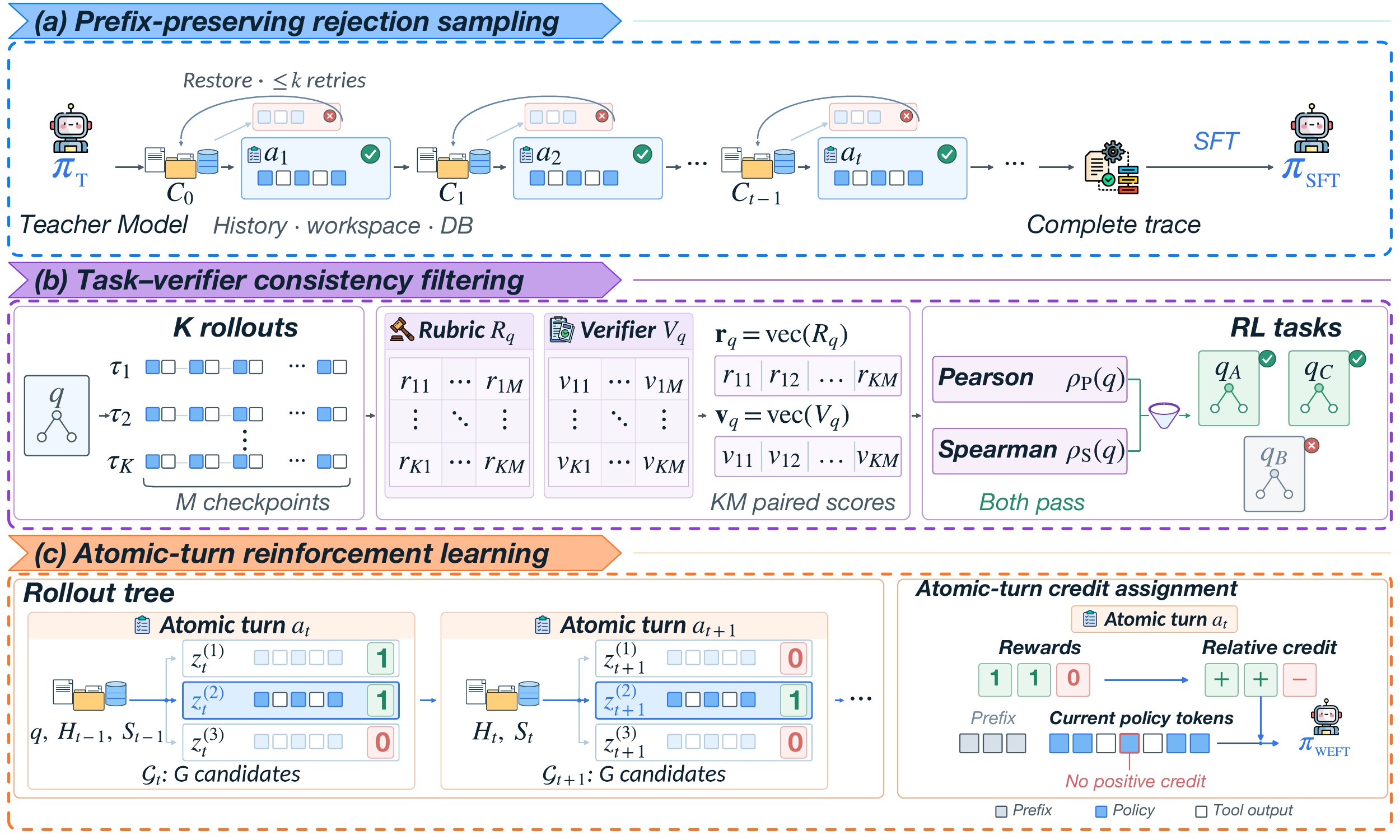}
    \par\nointerlineskip
    \caption{Post-training in \weft{}. (a) Verified prefixes form SFT
    trajectories. (b) RL task filtering compares rubric-based and executable
    scores. (c) Atomic-turn RL assigns local credit, masking positive credit
    for detected undesirable patterns.}
    \label{fig:weft-training}
    \vspace{-3mm}
\end{figure}

Construction and self-evolution provide the environments and tasks for
training, while stable post-training also requires high-quality training
data and reliable learning signals. \weft{} uses prefix-preserving rejection
sampling to construct supervised fine-tuning data and combines task
filtering with fine-grained rewards to provide reliable, stable learning
signals for reinforcement learning (Figure~\ref{fig:weft-training}).
MegaMCP provides scalable rollout support for both training stages.

\subsubsection{Prefix-Preserving Rejection Sampling}
\label{sec:prefix-preserving-rejection-sampling}

Rejecting complete trajectories after local failures discards successful
progress. We use atomic-task turns as rejection units: the teacher executes
tasks in dependency order, each checked by its verifier. Successful traces
and database and workspace states are retained for subsequent tasks. On failure,
we discard the attempt, restore pre-turn history and state, and resample only
that task, with up to $k$ additional retries. Once all tasks pass, retained
segments form a complete trajectory for SFT.

\subsubsection{Reinforcement Learning with Atomic-Turn Credit Assignment}
\label{sec:rl-from-verifiable-task-graphs}

\parhead{Task and verifier consistency filtering.}
\label{sec:consistency-guided-task-verifier-selection}
Verifiers may reject valid solutions or reward incomplete executions.
Before RL, a rubric-guided LLM judge independently scores each task's pilot
rollouts. We retain tasks only when both Pearson and Spearman correlations
with executable scores meet selection criteria. RL uses only executable
rewards.

\parhead{Atomic-turn group-relative credit assignment.}
\label{sec:atomic-turn-group-relative-credit}
Rollouts follow a fixed topological order of the task graph, with one atomic
task per turn. At turn $t$, a frozen rollout policy independently samples $G$ segments
from the same history in isolated, identical states restored by MegaMCP.
Each segment may contain multiple tool calls. The atomic-task verifier gives
$r_t^{(i)}=1$ on completion and $0$ otherwise. A uniformly selected passing
segment (or failed segment if all fail) supplies history and state for the
next turn. All candidates contribute to training.

We normalize rewards within each atomic-task group:
\begin{equation}
    A_t^{(i)}
    =
    \frac{r_t^{(i)}-\bar r_t}
    {\operatorname{sd}(\mathbf{r}_t)+\epsilon},
    \qquad
    \bar r_t=\frac{1}{G}\sum_{i=1}^{G}r_t^{(i)},
    \label{eq:atomic-turn-group-advantage}
\end{equation}
where $\mathbf{r}_t$ contains group rewards and $\epsilon>0$ stabilizes
normalization. Clipped GRPO applies these advantages only to current-turn
policy tokens, excluding prefixes, with token-mean aggregation across groups.
Equal rewards give zero advantages, leaving only KL and entropy regularization.

\begin{figure}[t]
    \centering
    \includegraphics[width=\linewidth]{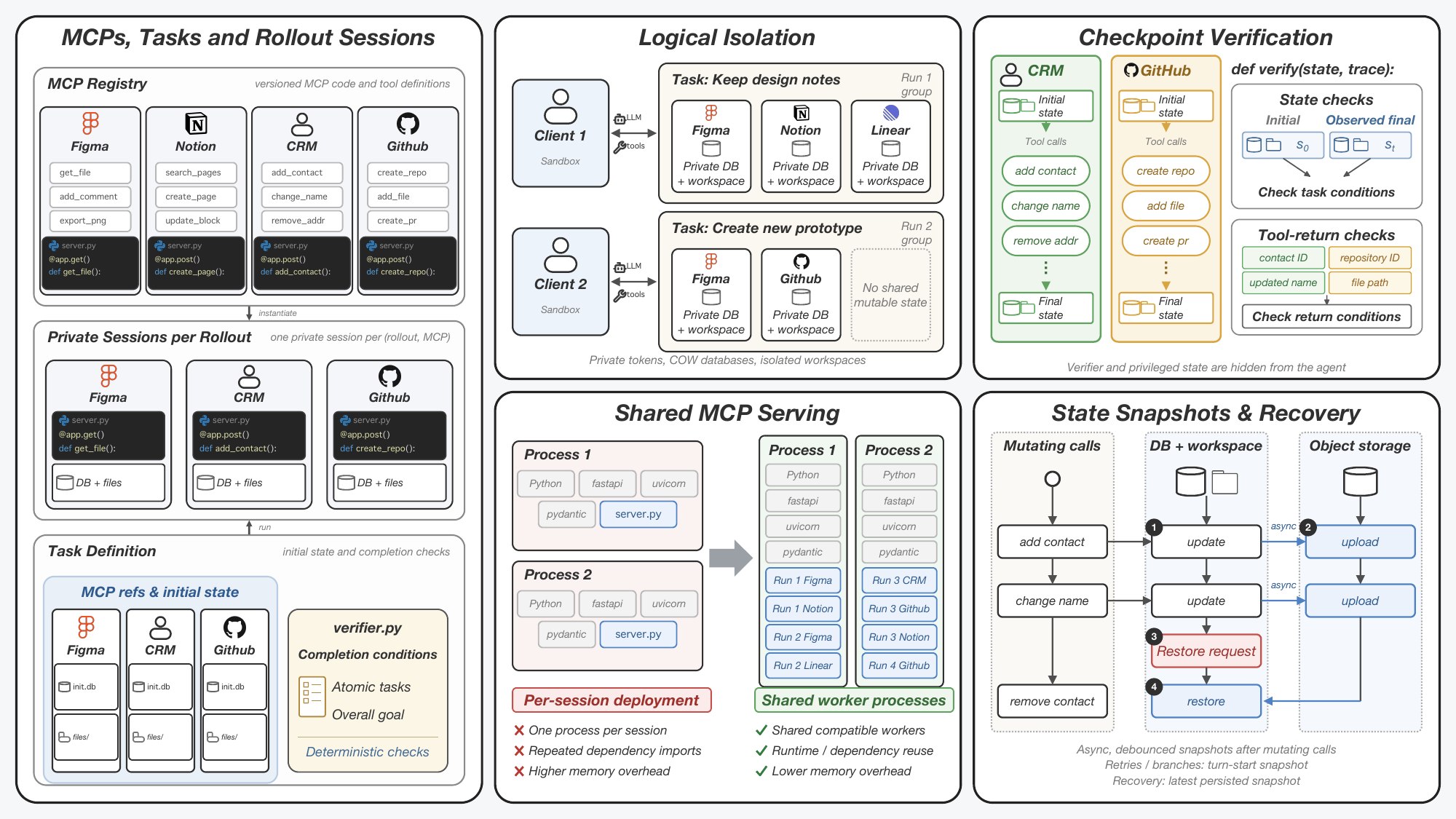}
    \caption{Overview of MegaMCP. A task's constituent MCPs are instantiated
    as private sessions within a unified lifecycle group. Compatible sessions
    share Python processes while retaining isolated mutable state. Initial
    and final databases and the complete tool-call ledger provide verification
    evidence, while persistent checkpoints support recovery and reloading.}
    \label{fig:megamcp-overview}
\end{figure}

Deterministic checks flag no-progress repetition, malformed tool-call or
reasoning structures, and interface violations, even in successful segments.
Flagged assistant turns receive token advantages $\min(A_t^{(i)},0)$,
removing only positive credit. Unflagged turns, including later recovery,
retain their advantages; KL and entropy regularization remain unchanged.

\subsubsection{MegaMCP: Shared Serving for Isolated Rollouts}
\label{sec:serving-mcps-during-rollout-with-megamcp}
\label{app:megamcp-runtime}

Prefix-preserving sampling and atomic-turn RL require large-scale concurrent
execution, along with local retries and candidate sampling from identical
states. Deploying MCPs separately in each rollout's sandbox requires repeated
provisioning of server code, initial databases, and workspaces, while incurring
per-rollout process startup and memory overhead. MegaMCP therefore separates reusable tool services from
each rollout's mutable state (Figure~\ref{fig:megamcp-overview}). It hosts MCP
services outside agent sandboxes and shares worker processes among compatible
sessions, reducing repeated deployment, startup costs, and memory use. Each
rollout retains private database and workspace state, with snapshots
supporting state restoration. Local retries and candidate sampling can thus
start from identical states without replaying previously completed
interactions.

\parhead{Registry and process sharing.}
The registry stores MCP metadata, source code, and tool definitions as
content-addressed records with monotonically increasing versions. Static
analysis screens for process-global side effects and assigns a load class
that limits how many instances share a process. Other servers use stronger
isolation tiers. A prefork master preloads common web and database
dependencies, then creates workers on demand. Each worker imports MCP
servers as separate modules and binds requests to the requesting session's
private database.

\parhead{Private state and session lifecycle.}
A task references its MCPs by registry identifier and its initial databases
by object-storage location, optionally including an executable verifier.
Each run creates a session group containing one private session per MCP.
Databases are copy-on-write clones of immutable initial snapshots, and
mutable workspaces are isolated across runs. Pausing or resuming a task
suspends or restores its MCP sessions and sandboxes as a group. Agents access
server-side state through MCP tools; internal database files, server
workspace paths, and credentials remain hidden. Each session has an
individual access token.

\begin{figure}[t]
  \centering
  \includegraphics[width=\linewidth]{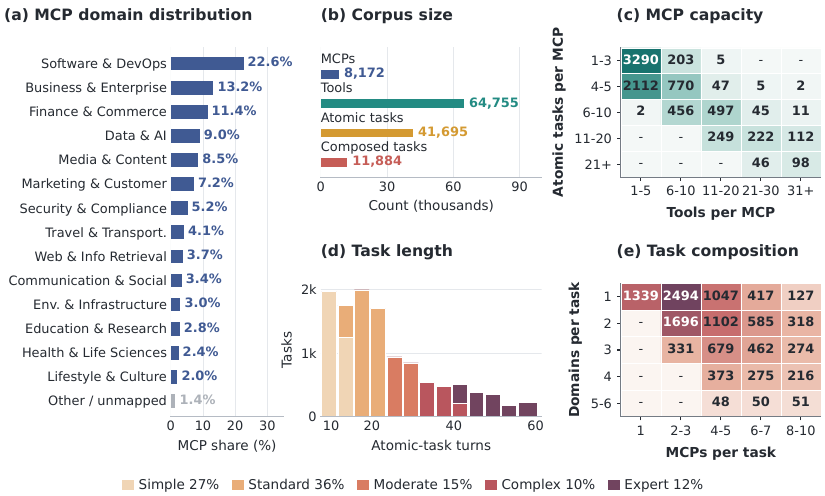}
  \caption{\textbf{Scale and composition of tool environments and agent tasks.}
  (a) MCP distribution across macro domains.
  (b) Total numbers of MCPs, tools, certified atomic tasks, and composed tasks.
  (c) MCP counts by tool and atomic-task inventories.
  (d) Atomic-task turns per composed task, colored by difficulty; legend
  percentages give the difficulty mix.
  (e) Task counts by the numbers of MCPs and primary domains.
  Heatmap cells report counts in the corresponding bins.}
  \label{fig:environment-task-corpus}
  \label{fig:tool-environment-scale}
  \label{fig:agent-task-composition}
\end{figure}

\parhead{Idle-session management.}
Agent generation leaves tool services idle between calls, with a median
interval exceeding ten seconds. After a cooling window, MegaMCP offloads an
idle session group's runtime while retaining its sessions, credentials, and
workspaces. The next tool call reloads the group. Longer-idle groups are
checkpointed and evicted from local disk, releasing resources between
periods of active tool use.

\parhead{State recovery.}
Database and workspace snapshots support local retries for SFT and RL
candidate sampling from identical states. After mutating calls, debounced
asynchronous checkpoints persist consistent snapshots to object storage
under immutable, generation-fenced keys. Resumption and crash recovery load
the latest persisted snapshot rather than replaying the full interaction.
Workers monitor disk and memory pressure, stop accepting placements when
unhealthy, and migrate affected sessions to healthy workers. Generation
fencing prevents writes from workers that no longer own a session, and a
reaper completes interrupted shutdowns idempotently. Persistent snapshots
also support re-execution after environment or task revision.

\parhead{Verification and operational feedback.}
MegaMCP retains each session's initial database, final database after a
group-wide freeze, and full tool-call ledger with arguments, results, and
outcomes. The verifier runs within the serving layer, which directly supplies
these immutable artifacts for task verification and reward computation. Its
code and privileged state remain inaccessible to the agent, preventing reward
hacking through direct inspection or modification of verifier internals.
Per-tool error rates and crash statistics are
associated with the responsible MCP definitions, guiding server revalidation
or stronger isolation in the execution-driven self-evolution loop
(Section~\ref{sec:evaluation}).

\FloatBarrier
\begingroup
\setlength{\textfloatsep}{10pt plus 2pt minus 2pt}
\setcounter{topnumber}{1}

\section{Experiments}
\label{sec:experiments}

We evaluate whether \weft{} improves performance across tool-use benchmarks
and examine how these gains arise. Our analysis studies how to obtain more
training value from existing tasks, improve the interaction system through
execution feedback, and use local verification for sampling and credit
assignment. We then measure the serving costs of supporting these rollouts
with MegaMCP.

\begin{table}[t]
  \centering
  \caption{\textbf{Main results on agentic tool-use benchmarks.}
  Scores (\%) are grouped by total model parameters. Bold and underlined
  values mark the best and second-best reported scores in each group;
  dashes denote unreported results.}
  \label{tab:main_results}
  \setlength{\tabcolsep}{2.1pt}
  \renewcommand{\arraystretch}{1.08}
  \newcommand{\resultmodelicon}[1]{%
    \makebox[1.45em][c]{%
      \raisebox{\dimexpr.35em-.5\height\relax}{%
        \includegraphics[width=1.35em,height=.95em,keepaspectratio]{figure/icons/#1}%
      }%
    }\hspace{.18em}%
  }
  \resizebox{\textwidth}{!}{%
  \begin{tabular}{l*{13}{c}}
    \toprule
    \multirow{2}{*}{\textbf{Method}} &
    \multicolumn{8}{c}{\textbf{BFCL V4}} &
    \multicolumn{4}{c}{\textbf{$\tau^2$-Bench}} &
    \textbf{Claw-Eval} \\
    \cmidrule(lr){2-9}\cmidrule(lr){10-13}\cmidrule(l){14-14}
    & WebSearch & Memory & Multi-T. & No live & Live & Relev. & Irrelev. & \textbf{Avg.}
    & Retail & Telecom & Airline & \textbf{Avg.} & \textbf{Avg.} \\
    \midrule
    \multicolumn{14}{c}{\textbf{\textit{Small- and Medium-Scale Open-Source Models (7B--32B)}}} \\
    \midrule
    \resultmodelicon{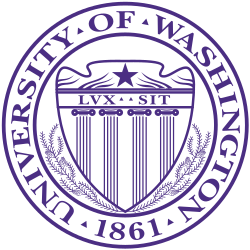}TOUCAN-7B
      & 21.0 & 18.5 & 17.8 & 81.0 & 73.9 & \underline{81.3} & 78.6 & 36.6
      & 22.8 & 10.5 & 20.0 & 17.7 & -- \\
    \resultmodelicon{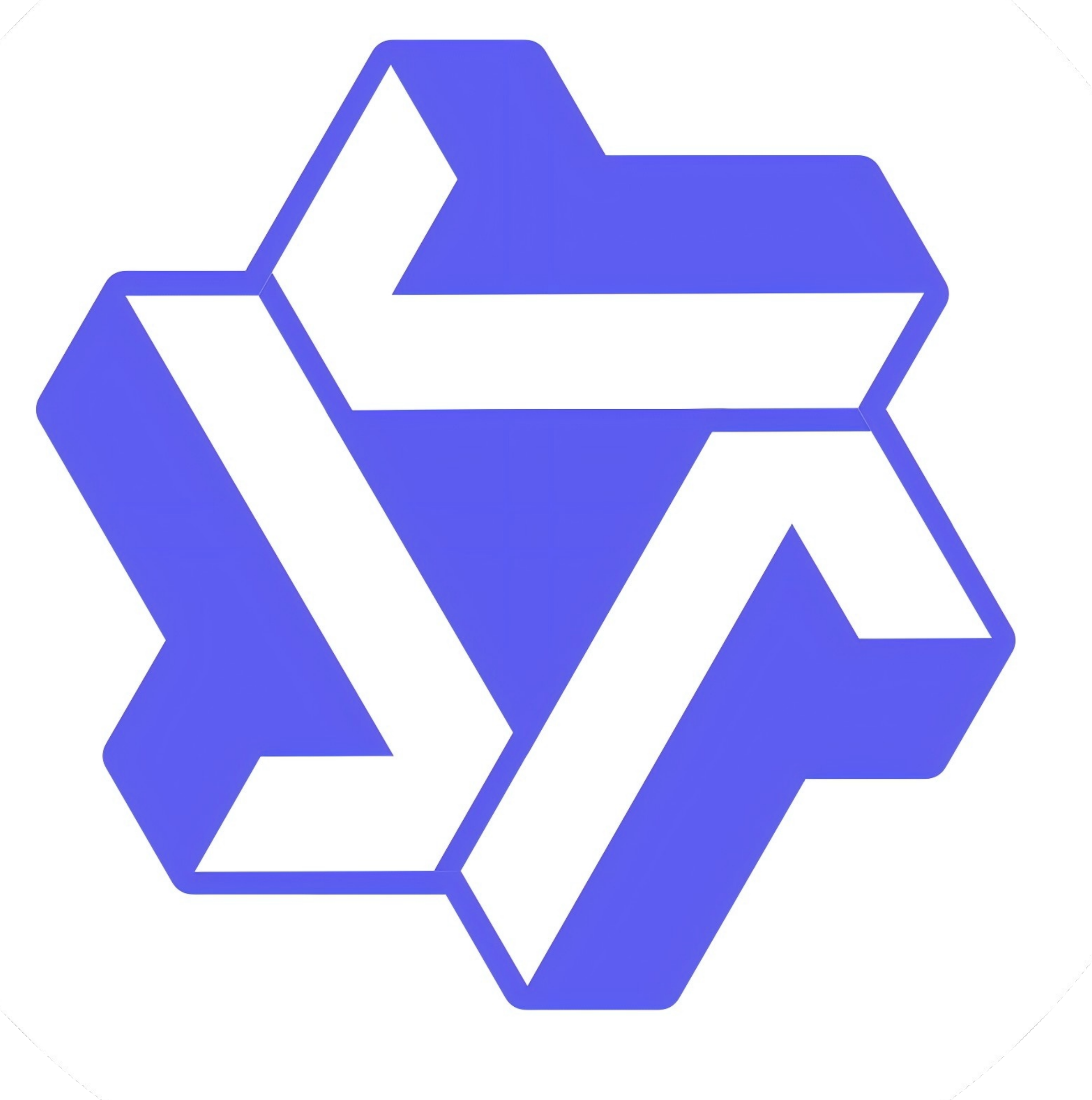}Qwen3-8B
      & 7.0 & 17.6 & 35.4 & \underline{90.2} & 80.9 & \underline{81.3} & 77.2 & 40.4
      & 34.0 & 18.0 & 26.5 & 26.2 & 25.6 \\
    \resultmodelicon{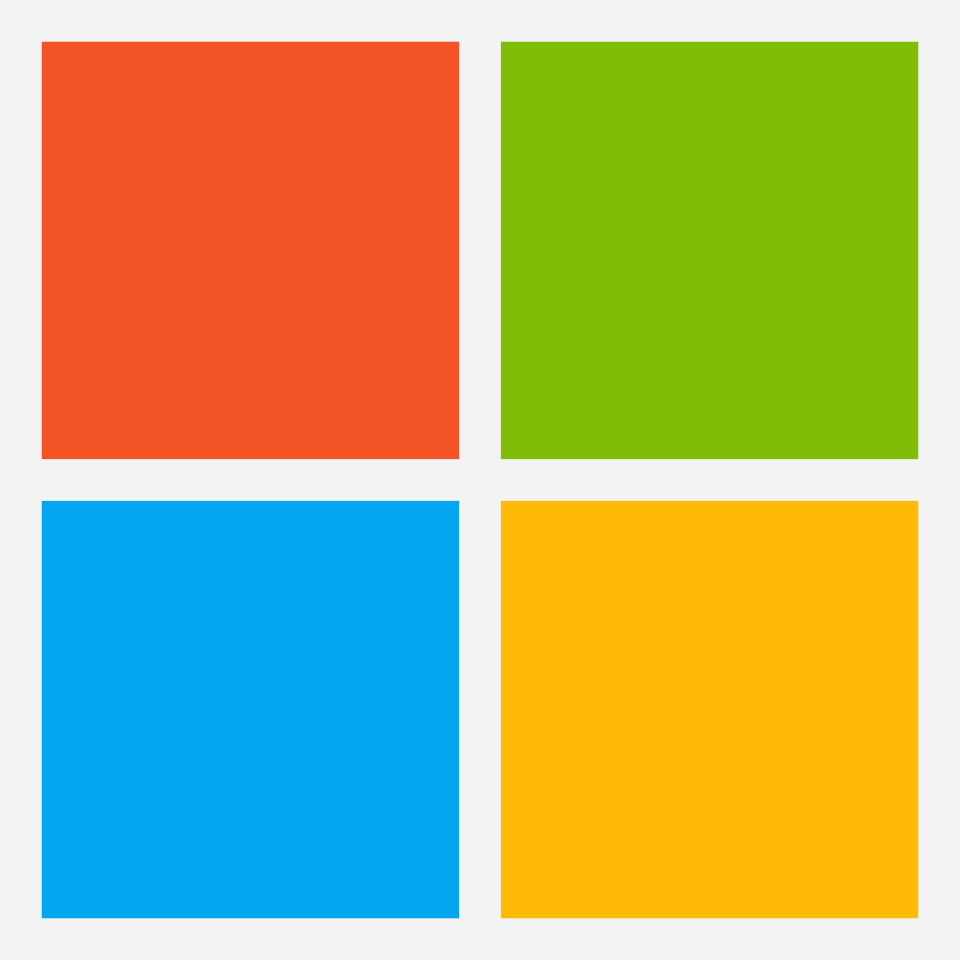}Simulator-8B
      & 17.5 & 6.0 & 4.1 & 47.6 & 44.6 & 31.3 & \textbf{87.3} & 23.9
      & 32.2 & 29.2 & 34.0 & 31.8 & -- \\
    \resultmodelicon{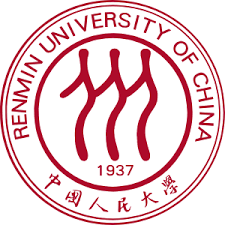}EnvScaler-8B
      & 23.0 & 21.9 & 47.1 & 88.5 & \underline{82.2} & \textbf{93.8} & 74.6 & 47.6
      & 49.6 & 32.7 & 31.5 & 37.9 & 22.6 \\
    \resultmodelicon{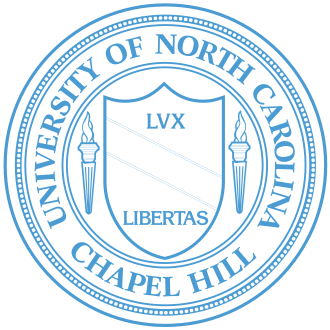}AWM-8B
      & 9.5 & 15.7 & 34.9 & \underline{90.2} & 80.5 & \textbf{93.8} & 73.9 & 40.0
      & 41.23 & 23.47 & 38.50 & 34.4 & 22.6 \\
    \resultmodelicon{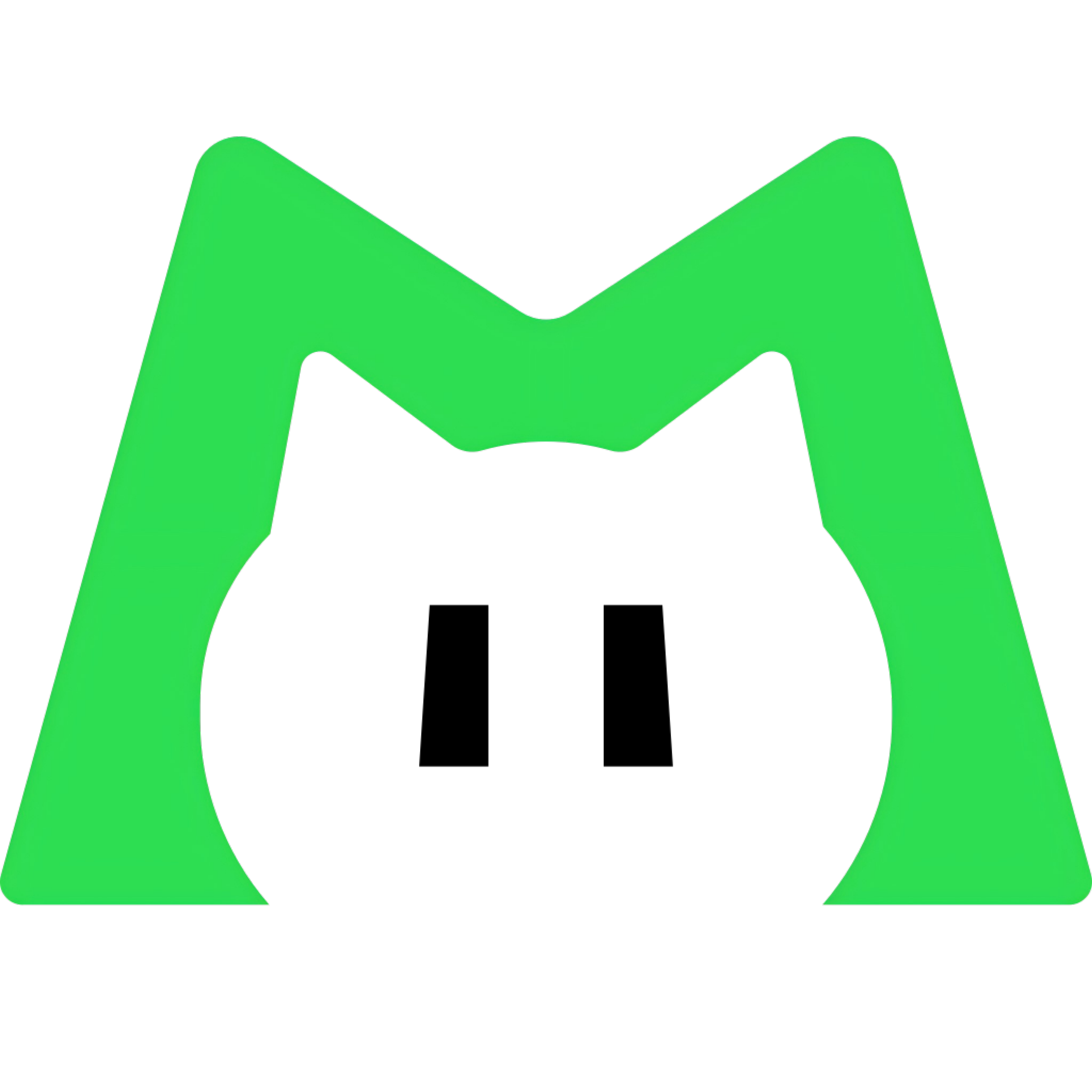}ScaleEnv-8B
      & -- & -- & -- & -- & -- & -- & -- & --
      & 50.9 & 27.2 & 37.5 & 38.5 & -- \\
    \resultmodelicon{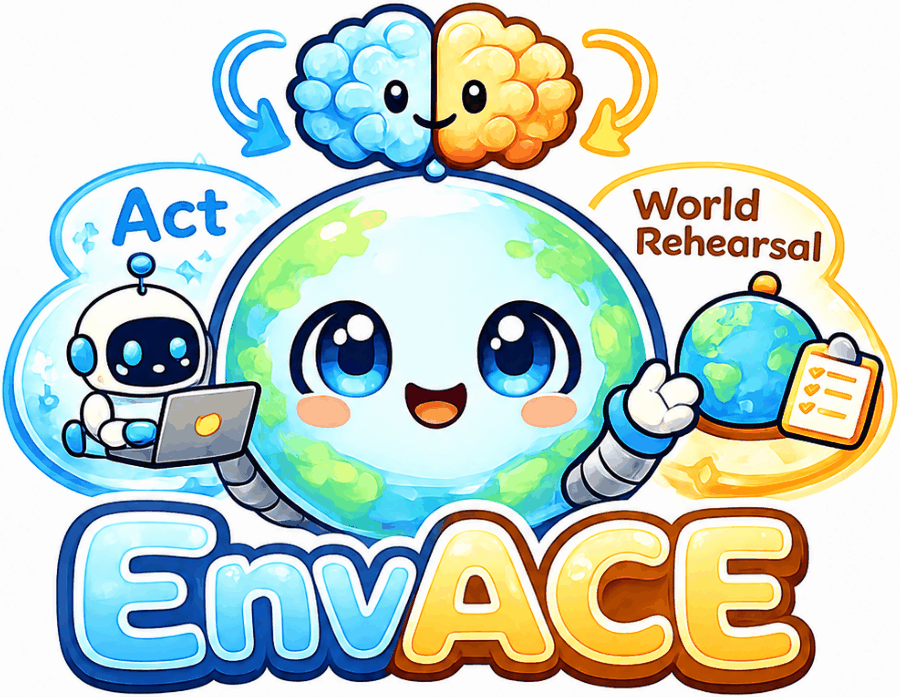}EnvACE-8B
      & 12.25 & 24.03 & 45.29 & 87.59 & 81.20 & -- & 83.19 & 46.04
      & 48.9 & 17.3 & 44.0 & 36.7 & -- \\
    \resultmodelicon{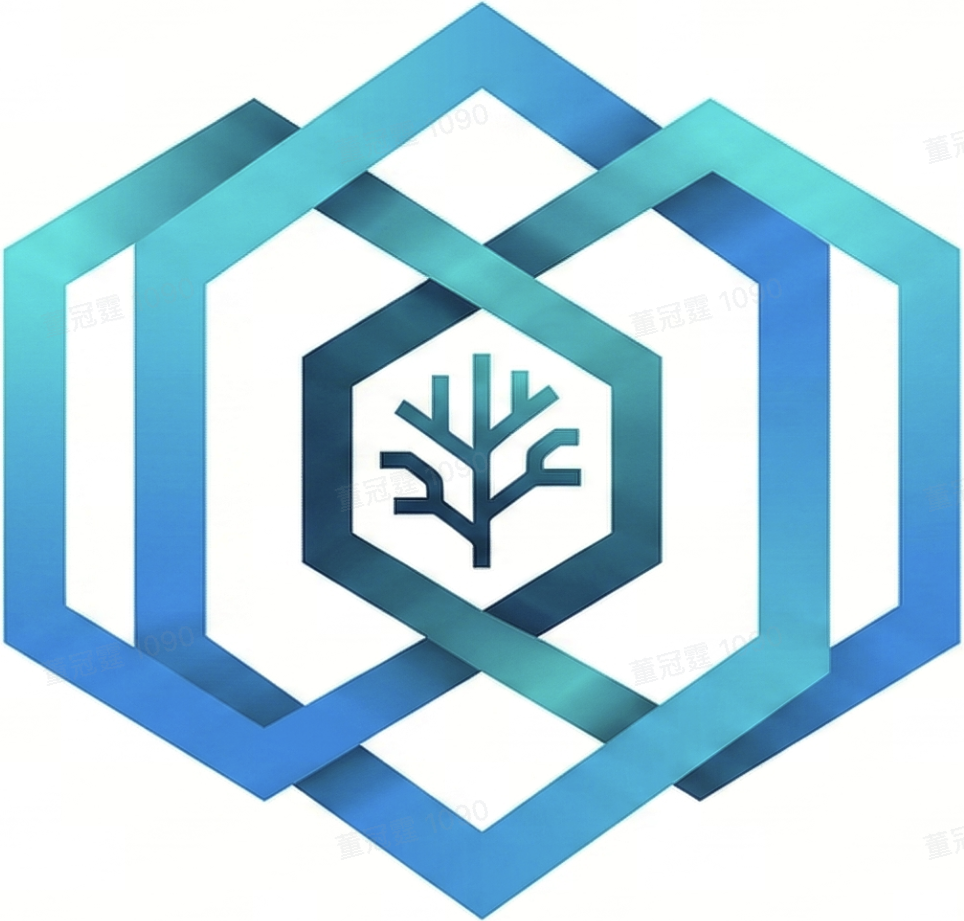}Agent-World-8B
      & 47.0 & 21.7 & 44.5 & 83.3 & 79.6 & \textbf{93.8} & 80.2 & 51.4
      & 72.8 & 50.9 & 40.0 & 61.8 & 30.5 \\
    \resultmodelicon{agent-world.png}Agent-World-14B
      & \underline{53.0} & 23.9 & \textbf{53.9} & 82.3 & 79.3 & \textbf{93.8} & 81.0 & \underline{55.8}
      & \underline{74.5} & 56.1 & \textbf{52.0} & \underline{65.4} & 31.5 \\
    \resultmodelicon{qwen.pdf}Qwen3-14B
      & 4.0 & 19.8 & 36.9 & 90.0 & \textbf{82.4} & \underline{81.3} & 79.4 & 41.0
      & 55.3 & 14.9 & 27.0 & 32.4 & 24.7 \\
    \resultmodelicon{awm.png}AWM-14B
      & 10.0 & 19.8 & 37.6 & \underline{90.2} & 81.5 & 75.0 & 79.4 & 42.4
      & 63.6 & 17.8 & 31.5 & 39.0 & 26.1 \\
    \resultmodelicon{qwen.pdf}Qwen3-32B
      & 26.0 & 15.7 & 43.3 & \textbf{90.3} & 82.0 & \underline{81.3} & 82.4 & 46.7
      & 59.5 & 27.2 & \underline{48.0} & 44.9 & -- \\
    \midrule
    \resultmodelicon{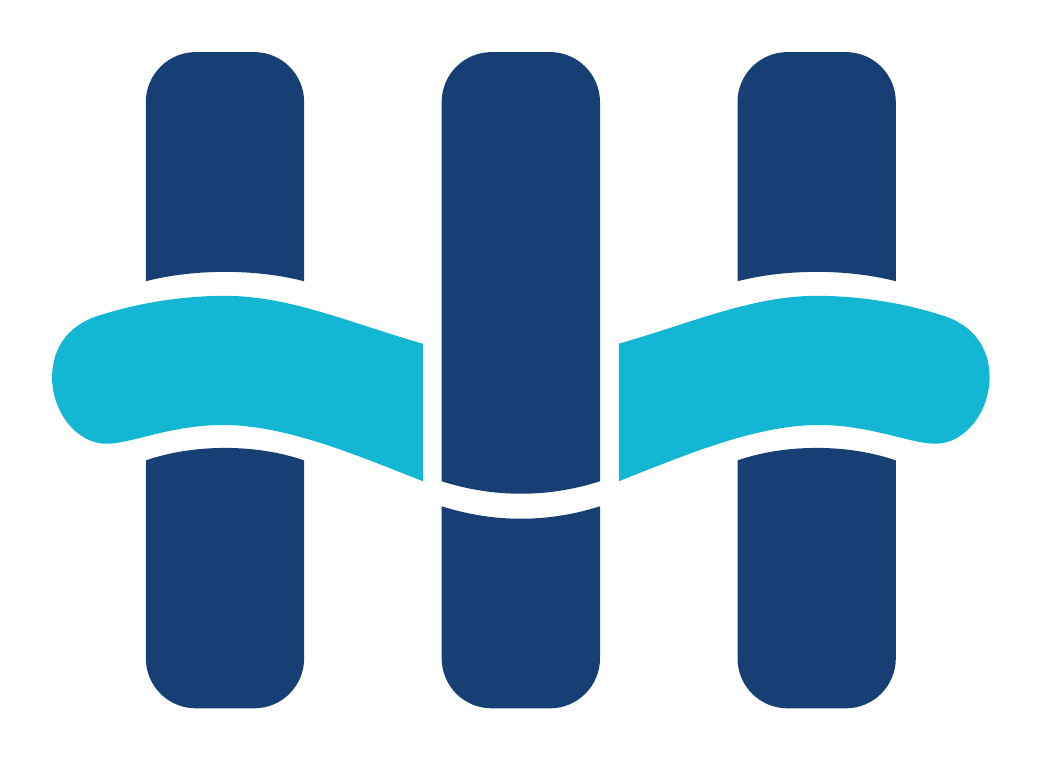}\textbf{\weft{}-8B}
      & 49.5 & \underline{33.12} & 47.75 & 77.9 & 69.5 & 81.25 & 66.88 & 52.28
      & 71.93 & \underline{72.81} & 42.0 & 62.25 & \underline{35.78} \\
    \resultmodelicon{../WEFT_logo_02B.pdf}\textbf{\weft{}-14B}
      & \textbf{68.00} & \textbf{49.25} & \underline{52.38} & 79.02 & 66.69 & 62.5 & \underline{84.75} & \textbf{62.21}
      & \textbf{74.56} & \textbf{76.32} & \textbf{52.0} & \textbf{67.63} & \textbf{43.77} \\
    \midrule
    \multicolumn{14}{c}{\textbf{\textit{Large-Scale Open-Source Models (35B--685B)}}} \\
    \midrule
    \resultmodelicon{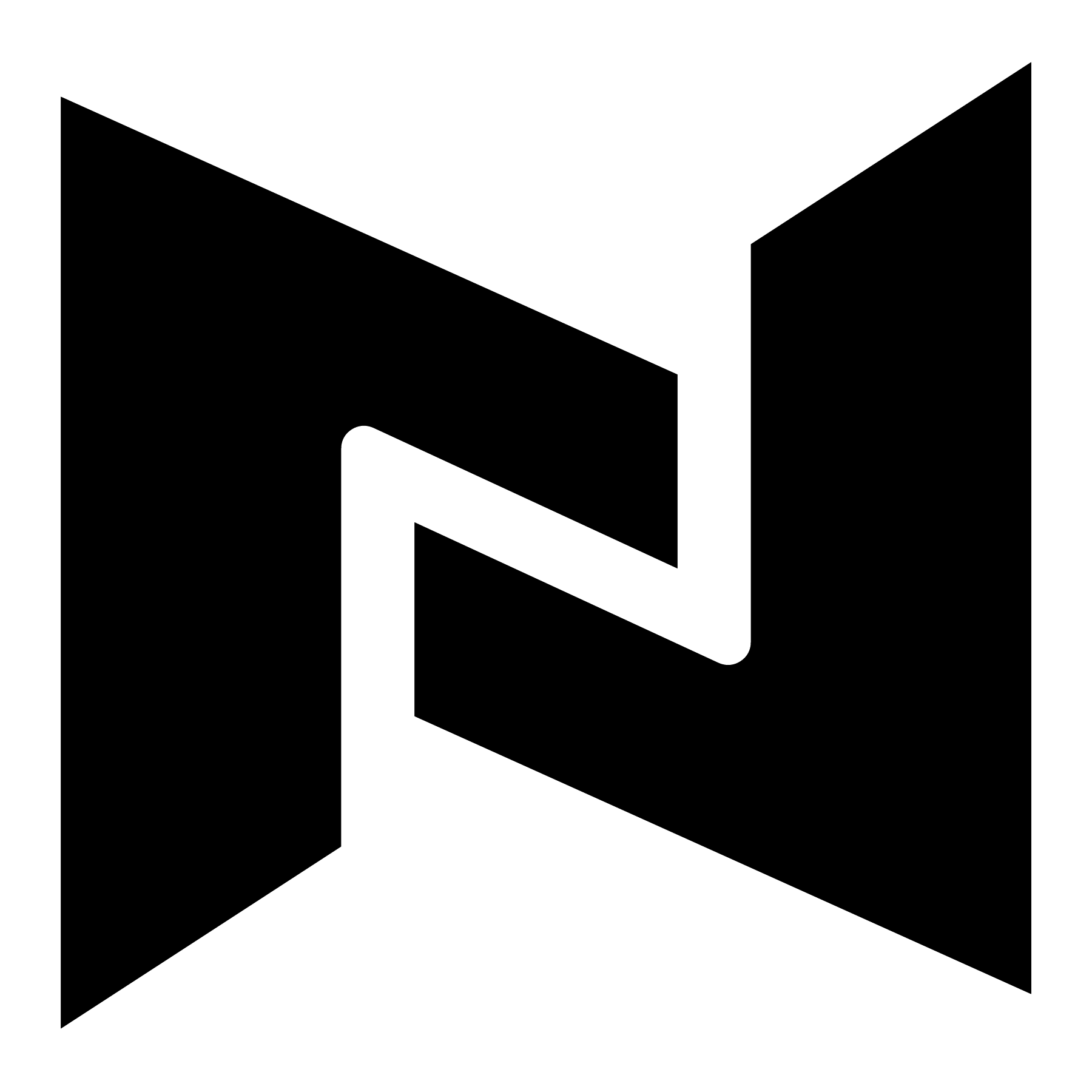}nex-n2-mini-35b
      & \textbf{72.50} & 39.14 & \textbf{62.12} & \underline{72.02} & 61.44 & 56.25 & \textbf{96.30} & \underline{63.94}
      & \underline{74.56} & 73.68 & \textbf{68.00} & 72.08 & \textbf{69.75} \\
    \resultmodelicon{qwen.pdf}Qwen3.5-35B-A3B
      & -- & -- & -- & -- & -- & -- & -- & \textbf{67.3}
      & -- & -- & -- & \textbf{81.2} & \underline{65.4} \\
    \resultmodelicon{qwen.pdf}Qwen3-235B-A22B
      & 54.0 & 23.9 & 45.4 & 37.4 & \underline{68.9} & \textbf{87.5} & 81.7 & 47.9
      & 71.9 & 58.0 & 45.6 & 58.5 & -- \\
    \resultmodelicon{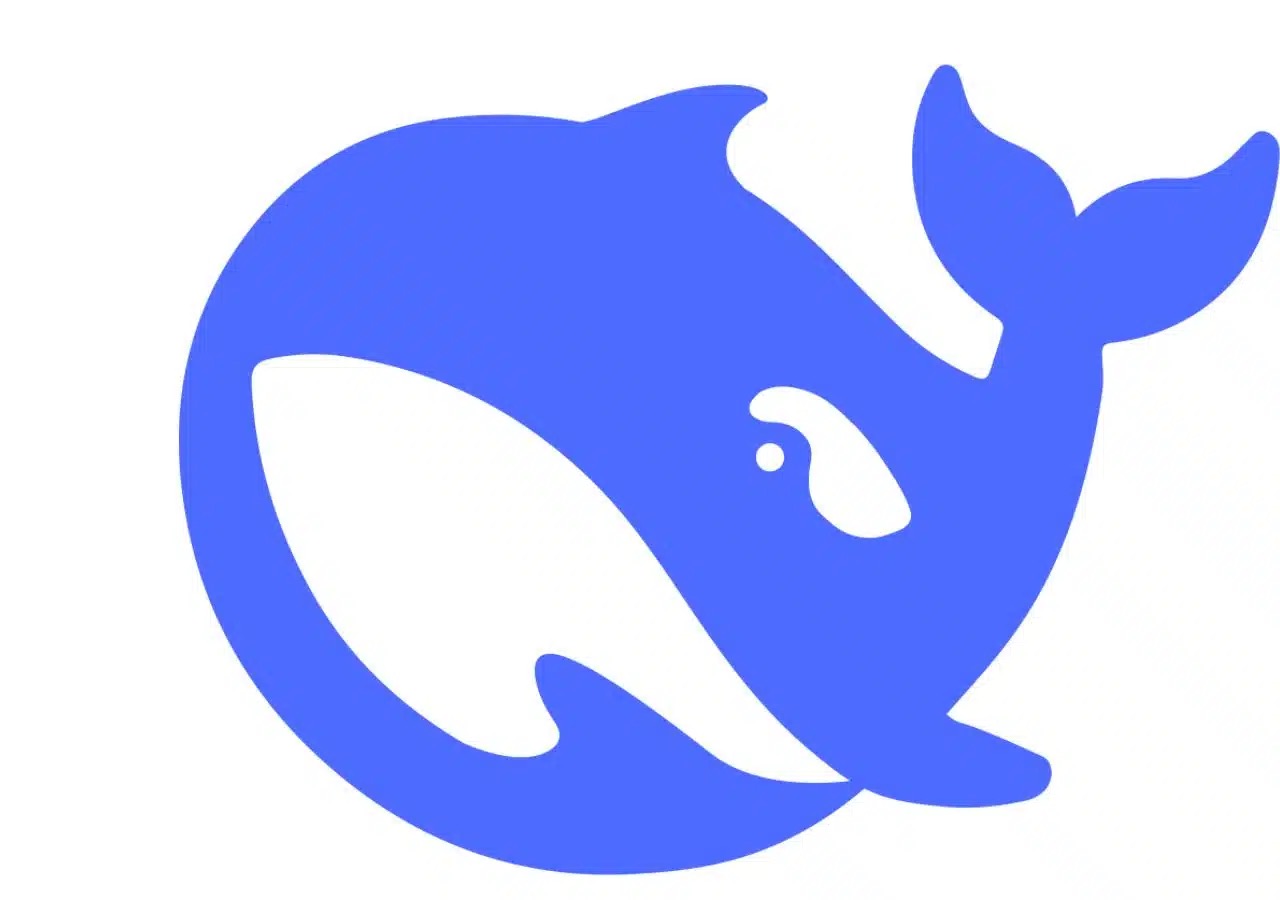}DeepSeek-V3.2-685B
      & \underline{69.5} & \underline{54.2} & 37.4 & 34.9 & 53.7 & 37.5 & \underline{93.2} & 54.1
      & \textbf{81.1} & \textbf{96.2} & 63.8 & \underline{80.3} & -- \\
    \midrule
    \resultmodelicon{../WEFT_logo_02B.pdf}\textbf{\weft{}-35B-A3B}
      & 55.0 & \textbf{66.02} & \underline{54.62} & \textbf{73.5} & \textbf{71.21} & \underline{68.75} & 83.39 & 63.4
      & 71.93 & \underline{85.09} & \underline{66.0} & 74.34 & 64.0 \\
    \bottomrule
  \end{tabular}%
  }
\end{table}

\subsection{Experimental Setup}
\label{sec:experimental-setup}

\parhead{Environment and task statistics.}
\label{sec:agent-task-scale-composition}
We construct 8,172 executable MCPs with 64,755 tools and 41,695 certified
atomic tasks. Figure~\ref{fig:tool-environment-scale}(a--c) shows domain
coverage and per-MCP capacity. Software \& DevOps is the largest domain,
accounting for 22.6\% of MCPs. Larger tool inventories tend to support
larger atomic-task libraries.
The 11,884 composed tasks have a median length of 20 atomic-task turns
(Figure~\ref{fig:agent-task-composition}(d--e)). Most tasks span multiple MCPs,
and more than half involve multiple domains. Five construction-time difficulty
profiles, Simple, Standard, Moderate, Complex, and Expert, specify increasing
task length and breadth across MCPs and domains. These tasks require sustained
cross-MCP interaction. Retaining intermediate progress and state continuity
is therefore central to collecting their trajectories.

\parhead{Models and training data.}
We post-train Qwen3-8B, Qwen3-14B, and Qwen3.5-35B-A3B from their base
models. We use \mbox{Kimi K3}~\citep{kimiteam2026kimik3openfrontier} to generate
supervised fine-tuning (SFT) trajectories for all 11,884 certified composed
tasks. For reinforcement
learning (RL), we select 1,954 of 5,000 candidate tasks using task and verifier
consistency filtering (Section~\ref{sec:consistency-guided-task-verifier-selection}).

\parhead{Evaluation.}
The main comparison and RL study use BFCL V4~\citep{patil2025bfcl},
$\tau^2$-Bench~\citep{barres2025tau2bench}, and
Claw-Eval~\citep{ye2026claweval}.  The data-construction ablations use
Toolathlon-Verified~\citep{li2025toolathlon},
AutomationBench~\citep{shepard2026automationbench}, and Claw-Eval.  Evaluation
follows each benchmark's native harness and official aggregate metric.

\FloatBarrier
\subsection{Main Results}
\label{sec:main-results}

\weft{}-8B and \weft{}-14B outperform all evaluated matched-size
environment-scaling baselines on BFCL V4, $\tau^2$-Bench, and Claw-Eval
(Table~\ref{tab:main_results}). \weft{}-14B achieves the highest aggregate
scores in the 7B--32B group, exceeding Agent-World-14B by 6.41, 2.23, and
12.27 percentage points, respectively. \weft{}-8B also improves all three
scores over Agent-World-8B, by 0.88, 0.45, and 5.28 points.
\looseness=-1

The category breakdown reveals recurring strengths in BFCL Memory and
$\tau^2$-Bench Telecom. Relative to the corresponding Agent-World models,
\weft{}-8B gains 11.42 and 21.91 points on these two categories, while
\weft{}-14B gains 25.35 and 20.22 points. \weft{}-14B also improves BFCL
WebSearch by 15.00 points, reaching 68.00\%. These strengths persist in the
larger model:
\weft{}-35B-A3B scores 66.02\% on Memory, the highest reported score
in the 35B--685B group and 11.82 points above the next-best result.
Its Telecom score of 85.09\% ranks second behind DeepSeek-V3.2 (96.2\%)
and exceeds nex-n2-mini-35b by 11.41 points. Memory and Telecom therefore
emerge as recurring strengths across the evaluated model sizes, rather than
isolated high scores at a single scale.

On the advanced benchmarks Toolathlon-Verified and
AutomationBench, \weft{}-35B-A3B achieves 45.99\% and 26.50\%, respectively
(Table~\ref{tab:harness-scaling}), extending the evaluation beyond the main
benchmarks to more demanding multi-tool workflows.

\Needspace{10\baselineskip}
\subsection{Understanding the Post-Training Gains}
\label{sec:understanding-post-training-gains}

\paragraph{Task Views and Harness Scaling}
We test whether a fixed set of tasks can provide more useful training
experience through different forms of interaction
(Table~\ref{tab:harness-scaling}). Under ReAct, SimUser alone scores below
Agentic on all three benchmarks, yet combining the two outperforms either
view, averaging 3.18 points over Agentic. Thus, a view that is weaker in
isolation can still improve the training mixture.
Adding OpenClaw and Hermes with the view mixture fixed yields a further
9.71-point mean gain: 13.58 on Toolathlon-Verified, 12.83 on AutomationBench,
and 2.71 on Claw-Eval. All gains are measured under benchmark-native harnesses.
Thus, a fixed task set can support richer training through multiple disclosure
views and execution frameworks. This provides a way to expand useful training
experience when new executable, verifiable tasks are difficult to obtain.

\begingroup
\setlength{\intextsep}{4pt plus 2pt minus 2pt}
\begin{table}[!htb]
  \centering
  \small
  \setlength{\parskip}{0pt}
  \setlength{\abovecaptionskip}{6pt}
  \setlength{\tabcolsep}{5pt}
  \renewcommand{\arraystretch}{1.2}
  \newlength{\hermescolumnwidth}
  \settowidth{\hermescolumnwidth}{Hermes}

  \caption{
    \textbf{Task views and harness scaling.}
    Benchmark scores (\%) of \weft{}-35B-A3B under different training-data
    configurations with the task set fixed.
  }
  \label{tab:harness-scaling}

  \begin{tabularx}{\linewidth}{@{}>{\raggedright\arraybackslash}X*{5}{r}@{\hspace{16pt}}>{\columncolor{white}[0pt][0pt]\centering\arraybackslash}p{\hermescolumnwidth}@{}}
    \toprule[0.8pt]
    & \multicolumn{3}{c}{\textbf{Disclosure views}}
      & \multicolumn{3}{c}{\textbf{Harness scaling (cumulative)}} \\
    \cmidrule(lr){2-4}\cmidrule(l){5-7}
    \textbf{Benchmark}
      & Agentic & SimUser & Both
      & ReAct & + OpenClaw & \makebox[0pt][r]{+\ }Hermes \\
    \midrule[0.4pt]
    Toolathlon-Verified
      & 27.78 & 19.44 & \cellcolor[HTML]{EDF3FA}\textbf{32.41}
      & 32.41 & 39.51 & \cellcolor[HTML]{EDF3FA}\textbf{45.99} \\
    AutomationBench
      & 10.50 & 4.67 & \cellcolor[HTML]{EDF3FA}\textbf{13.67}
      & 13.67 & 22.00 & \cellcolor[HTML]{EDF3FA}\textbf{26.50} \\
    Claw-Eval
      & 59.55 & 57.51 & \cellcolor[HTML]{EDF3FA}\textbf{61.29}
      & 61.29 & 63.58 & \cellcolor[HTML]{EDF3FA}\textbf{64.00} \\
    \bottomrule[0.8pt]
  \end{tabularx}
\end{table}
\endgroup

\Needspace{17\baselineskip}
\begin{wrapfigure}{r}{0.44\linewidth}
  \vspace{-8pt}
  \centering
  \setlength{\abovecaptionskip}{4pt}
  \includegraphics[width=\linewidth]{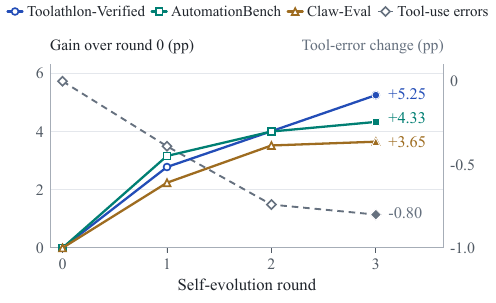}\par
  \caption{\textbf{Self-evolution across rounds.}}
  \label{fig:self-evolution}
  \vspace{-8pt}
\end{wrapfigure}

\paragraph{Execution-Driven Self-Evolution}
Execution traces can also guide improvements to the interaction system
that produces them. With the task set and rollout budget fixed, we compare
the initial construction ($r=0$) with up to three self-evolution rounds
(Figure~\ref{fig:self-evolution}). Tool-error rate measures the fraction of
tool calls returning errors in selected teacher trajectories, including
errors from which the teacher recovers.
Three rounds reduce tool-error rate from 1.76\% to 0.96\% (45.5\% relative)
and improve Toolathlon-Verified, AutomationBench, and Claw-Eval by 5.25,
4.33, and 3.65 points. The first two rounds capture most of the gains on
AutomationBench and Claw-Eval, while Toolathlon-Verified continues to improve
in the third round as tool errors decline monotonically. The reduction in
errors within selected trajectories shows that outcome filtering still
leaves room to improve the underlying interactions. Alongside the downstream
gains, this supports using execution experience both as policy-training data
and as evidence for revising the system that generates it.

\paragraph{Prefix-Preserving Rejection Sampling}
Figure~\ref{fig:rejection-sampling} compares prefix-preserving and end-to-end
sampling with the task pool, verifier, and accepted-trace count fixed.
Each allows $k\in\{1,2,3\}$ additional retries per atomic-task turn or
full trajectory, respectively.
At each retry limit $k$, prefix-preserving sampling outperforms end-to-end
sampling. At $k=3$, it gains 1.85 points on Toolathlon-Verified and 2.17 on
AutomationBench, with Claw-Eval nearly tied. Since the accepted-trace count
is fixed, these gains concern the training value of the resulting SFT data,
not an increase in the number of accepted examples. For tasks with verifiable
and recoverable intermediate states, the results support using local task
boundaries as rejection units while retaining completed progress.

\begin{figure}[!htb]
  \centering
  \setlength{\parskip}{0pt}
  \setlength{\abovecaptionskip}{4pt}
  \includegraphics[width=\linewidth,trim=0 7bp 0 3bp,clip]{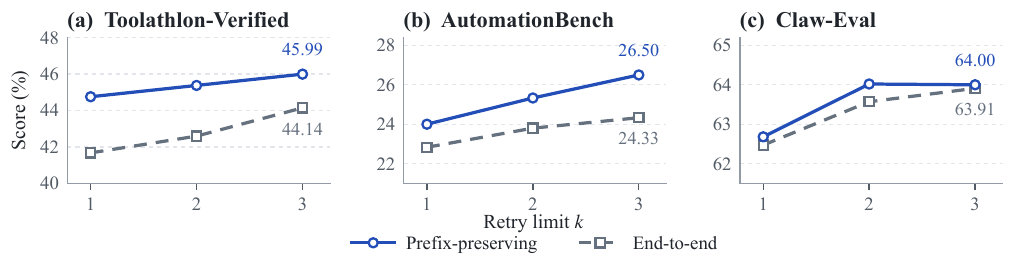}\par
  \caption{\textbf{Rejection sampling across retry limits.}
  Downstream scores of \weft{}-35B-A3B.}
  \label{fig:rejection-sampling}
\end{figure}

\paragraph{Reinforcement Learning and Credit Assignment}

RL gains depend on how verified outcomes are assigned to policy updates.
We compare trajectory-level and atomic-turn GRPO with SFT initialization,
task pool, reward verification, and positive-credit masking fixed
(Table~\ref{tab:rl-gains}; Appendix~\ref{app:credit-assignment}).
Trajectory-level GRPO improves BFCL V4 and Claw-Eval but decreases
$\tau^2$-Bench scores relative to SFT by 3.17 and 0.84 points for 8B and
14B, respectively. Atomic-turn GRPO instead improves all three benchmarks
at both sizes, exceeding the trajectory-level variant by 5.69 and 4.43
points on $\tau^2$-Bench.
The largest atomic-turn gains over SFT occur on Claw-Eval: 4.59 points for
8B and 8.54 for 14B. Thus, local credit assignment both improves on SFT
and avoids the $\tau^2$-Bench regression observed with trajectory-level
updates. This contrast supports assigning each segment credit for its own
task outcome, rather than applying an advantage derived from average
completion over the full trajectory to every turn. Local completion checks
can therefore serve as both evaluation criteria and the basis for assigning
learning signals to the relevant policy decisions.

\begingroup
\setlength{\intextsep}{4pt plus 2pt minus 2pt}
\begin{table}[!htb]
  \centering
  \small
  \setlength{\parskip}{0pt}
  \setlength{\abovecaptionskip}{4pt}
  \setlength{\tabcolsep}{3pt}
  \caption{\textbf{RL gains and credit assignment.} Benchmark scores (\%) after SFT and RL.}
  \label{tab:rl-gains}
  \begin{tabularx}{\linewidth}{@{}l*{6}{>{\centering\arraybackslash}X}@{}}
    \toprule
    \multirow{2}{*}{Training configuration}
      & \multicolumn{3}{c}{\weft{}-8B}
      & \multicolumn{3}{c}{\weft{}-14B} \\
    \cmidrule(lr){2-4}\cmidrule(l){5-7}
      & BFCL V4 & $\tau^2$-Bench & Claw-Eval
      & BFCL V4 & $\tau^2$-Bench & Claw-Eval \\
    \midrule
    SFT
      & 52.02 & 59.73 & 31.19 & 59.69 & 64.04 & 35.23 \\
    \quad + RL (trajectory-level GRPO)
      & 52.25 & 56.56 & 34.92 & 60.47 & 63.20 & 41.64 \\
    \rowcolor[HTML]{EDF3FA}
    \quad + RL (atomic-turn GRPO)
      & \textbf{52.28} & \textbf{62.25} & \textbf{35.78}
      & \textbf{62.21} & \textbf{67.63} & \textbf{43.77} \\
    \bottomrule
  \end{tabularx}
\end{table}
\endgroup

\paragraph{RL Training Curves}
Figure~\ref{fig:rl-learning-weft-14b} complements the final benchmark scores
with the optimization history of \weft{}-14B. Mean trajectory reward
increases overall while policy entropy declines. The reported reward averages
binary checkpoint outcomes over the full trajectory, tracking task completion
across the workflow; training advantages use the current atomic task's binary
reward. Appendix~\ref{app:learning-dynamics} details trajectory reconstruction
and reward aggregation.

\begin{figure}[!htb]
  \centering
  \includegraphics[width=0.85\linewidth]{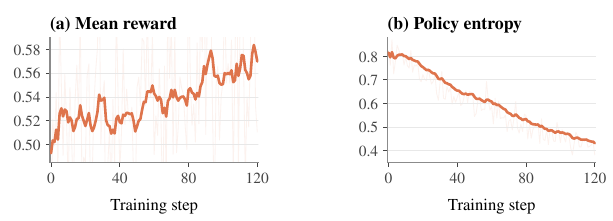}
  \caption{\textbf{RL training curves of \weft{}-14B.}
  (a) Mean trajectory reward; (b) policy entropy.}
  \label{fig:rl-learning-weft-14b}
\end{figure}

\paragraph{MegaMCP at Rollout Scale}
We compare shared and sandbox-local MCP serving on identical task traces
to measure the cost of repeatedly deploying tool services
(Figure~\ref{fig:megamcp-summary}). At 1,000 tasks, MegaMCP reduces total
sandbox upload, including runner files, from 773.5 to 34.8\,MiB (95.5\%)
by removing repeated delivery of MCP source and initialized databases.
Sharing processes also reduces resident memory by 77.6\% for 52 instances
in 8 processes and by 95.7\% for 50 sessions in one process, relative to
one process per session. These reductions show that independent rollout
state need not require a separate deployment of the full tool-service stack
for every rollout.
With both systems cold at 100 tasks, median time to the first tool call
falls from 10.455 to 4.794\,s (54.1\%). Shared serving thus reduces the
cost of repeatedly initializing tool services while preserving isolated
rollout state. Appendix~\ref{app:megamcp-benchmark} details the timing
conditions and reports full latency measurements and historical 1,000-task
cold-repeat records.

\begingroup
\setlength{\intextsep}{4pt plus 2pt minus 2pt}
\begin{figure}[!htb]
  \centering
  \setlength{\parskip}{0pt}
  \setlength{\abovecaptionskip}{4pt}
  \includegraphics[width=\linewidth]{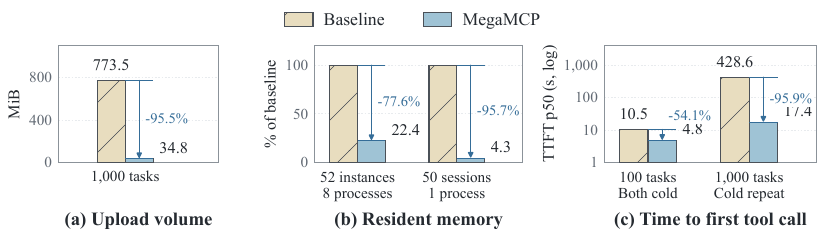}\par
  \caption{\textbf{MegaMCP resource use and startup latency.}
  (a) Total sandbox upload. (b) Memory relative to one process per session.
  (c) Median time to first tool call (log scale); 1,000-task bars show
  historical cold repeats. Local serving is the baseline in (a,c);
  conditions are in Appendix~\ref{app:megamcp-benchmark}.}
  \label{fig:megamcp-summary}
\end{figure}
\endgroup

\FloatBarrier
\setcounter{topnumber}{2}
\endgroup

\Needspace{13\baselineskip}
\section{Conclusion}
\label{sec:conclusion}

\weft{} shows that scaling tool-use post-training is fundamentally a systems
problem rather than a matter of simply increasing the number of environments.
Its central principle is to give execution experience a dual role: as data for
policy learning and as evidence for improving the interaction system that
produces it. Controlled experiments show that diversity across tasks, disclosure
views, and agent harnesses broadens the training distribution; execution-driven
self-evolution reduces tool-call errors and improves downstream performance;
and stable sampling, fine-grained credit assignment, and scalable rollout
infrastructure make these gains effective at scale. Overall, these findings
support system-level scaling for reliable tool-use agents.

\bibliography{iclr2027_conference}
\bibliographystyle{iclr2027_conference}

\clearpage
\appendix
\section{Additional Method Details}
\label{app:method-details}

\subsection{Atomic-Task Construction and Certification}
\label{app:weft-atomic-task-certification}

This section details the three checks used to construct atomic-task libraries
in Section~\ref{sec:task-world-composition}. Each task description records
the intent, required inputs, ordered tool chain, and completion conditions.

\parhead{E2E execution validation.}
Each test starts from an independently initialized environment state and
executes MCP tools. A task passes when returned observations and persistent
state satisfy its completion conditions. Passing tasks accumulate in $V$;
test, response, and state diagnostics from failures accumulate in $F$ for
repair. Passing tasks are retained without regeneration.

\parhead{Tool-chain orthogonality.}
Let $\ell_i$ denote the ordered tool-name sequence of task $a_i\in V$.
Two tasks are redundant when either chain is a subsequence of the other:
\[
    \ell_i\preceq\ell_j \quad\text{or}\quad \ell_j\preceq\ell_i.
\]
Here $\preceq$ preserves order but does not require contiguity, and includes
equality. For example, $(u_1,u_3)\preceq(u_1,u_2,u_3)$.
Filtering removes the task with the strictly contained chain; identical
chains retain the earlier task. Sharing tools alone does not imply
redundancy. Each round filters $V$ to obtain the retained set $A$ and
redundancy feedback $D$.

\parhead{Tool coverage.}
Coverage is the number of distinct MCP tools in the retained task chains
divided by the MCP's total number of tools. We compute it after orthogonality
filtering, using the declared tool chains of retained tasks. Let $T$ denote
the MCP tool set and $T_a$ the tools in task $a$'s chain. Uncovered tools guide
further generation toward the target coverage $\eta$.

\parhead{Incremental construction.}
Algorithm~\ref{alg:atomic-task-certification} combines these checks, using
failure, redundancy, and coverage feedback in the ongoing synthesis
conversation. The budget $R$ includes the initial validation round.
Construction stops when the target coverage is reached, the round budget
is exhausted, or no new candidates remain. We return the retained tasks and
attained coverage, and save the tasks, E2E tests, and validation reports.

\begin{algorithm}[H]
    \caption{Incremental Atomic-Task Construction and Certification}
    \label{alg:atomic-task-certification}
    \small
    \algrenewcommand\algorithmicrequire{\textbf{Input:}}
    \algrenewcommand\algorithmicensure{\textbf{Output:}}
    \begin{algorithmic}[1]
        \Require Certified MCP $m$ with tools $T$, state schema, and execution contracts;
        \Statex synthesis agent $g$, target coverage $\eta$, maximum validation rounds $R\geq 1$
        \Ensure Filtered task set $A$ and attained tool coverage $c$
        \State $V\gets\emptyset$; $F\gets\emptyset$
        \State $B\gets\Call{Synthesize}{g,m}$
        \For{$r=1,\ldots,R$}
            \State $(P,F_r)\gets\Call{ValidateE2E}{m,B}$ \Comment{Independent state per task}
            \State $V\gets V\cup P$; $F\gets F\cup F_r$
            \State $(A,D)\gets\Call{FilterOrthogonal}{V}$
            \State $C\gets\bigcup_{a\in A}T_a$
            \State $c\gets |C|/|T|$
            \If{$(A\neq\emptyset\ \text{and}\ c\geq\eta)\ \text{or}\ r=R$}
                \State \textbf{break}
            \EndIf
            \State $B\gets\Call{ReviseAndExpand}{g,m,V,F,D,T\setminus C}$
            \State Remove from $B$ tasks whose identifiers already occur in $V$
            \If{$B=\emptyset$}
                \State \textbf{break}
            \EndIf
        \EndFor
        \State \Return $(A,c)$
    \end{algorithmic}
\end{algorithm}

\subsection{Cross-MCP Task Composition}
\label{app:weft-task-sampling}

The atomic-task library is unevenly distributed across MCPs and domains.
To control the resulting task distribution, we use five difficulty profiles,
\emph{Simple}, \emph{Standard}, \emph{Moderate}, \emph{Complex}, and
\emph{Expert}. Each specifies ranges for atomic-task count, MCP count, and
domain breadth, together with a target corpus proportion. These dimensions
control task length, the number of services to coordinate, and the range of
business contexts involved.

Given the requested corpus size, largest-remainder allocation converts
profile proportions into integer quotas. Within a profile's constraints,
the sampler uses atomic-task capacity and historical usage to select domains
and MCPs, favoring capabilities underrepresented in generated tasks. It ranks
candidate combinations by shared entities, complementary operations, and
functional redundancy. The profile mixture controls corpus-level difficulty,
while coverage-aware selection broadens the capabilities represented within
each profile.

Composition requires a surplus of candidate atomic tasks because not every
local objective fits a shared scenario. For a selected MCP set $M$ and target
atomic-task count $k$, we require
\[
    \sum_{m\in M}c_m\ge\left\lceil\rho k\right\rceil,
    \qquad \rho>1,
\]
where $c_m$ is the number of certified atomic tasks in MCP $m$ and $\rho$
provides a proportional capacity surplus. A semantic evaluation model checks
whether the selected MCPs support a coherent shared objective. Accepted
combinations count toward the profile quota and update usage
statistics before an atomic-task set is sampled. Rejected
combinations are resampled within the same profile.

\FloatBarrier
\Needspace{8\baselineskip}
\section{Training and Evaluation Details}
\label{app:experimental-details}

\subsection{Training Configuration}
\label{app:training-configuration}

\parhead{SFT sampling.}
\mbox{Kimi K3}~\citep{kimiteam2026kimik3openfrontier} generates SFT trajectories
using prefix-preserving rejection
sampling with up to $k=3$ retries per atomic-task turn, allowing up to four
attempts including the initial attempt.

\parhead{RL task selection.}
We independently synthesize 5,000 candidate task--verifier pairs.  For each
task, \mbox{Nex-N2-mini} generates 16 pilot rollouts, which are independently
scored by a rubric-based LLM judge and an executable verifier.  We select
pairs based on consistency between the two sets of scores, following
Section~\ref{sec:consistency-guided-task-verifier-selection}. The resulting
1,954 tasks are used for both the 8B and 14B models.

\parhead{RL optimization.}
The 8B and 14B models use atomic-turn GRPO with the optimization and rollout
settings in Table~\ref{tab:rl-training-configuration}.
The reference policy is fixed at each model's RL initialization checkpoint.

\begin{table}[!htb]
  \centering
  \small
  \caption{\textbf{RL training configuration.} Optimization and rollout settings
  for the 8B and 14B models.}
  \label{tab:rl-training-configuration}
  \renewcommand{\arraystretch}{1.12}
  \begin{tabularx}{\linewidth}{@{}>{\raggedright\arraybackslash}p{0.35\linewidth}
      >{\raggedright\arraybackslash}X@{}}
    \toprule
    \textbf{Configuration} & \textbf{Setting} \\
    \midrule
    Optimizer & Adam \\
    Learning rate and schedule & $10^{-6}$, constant \\
    Adam moments & $\beta_1=0.9$, $\beta_2=0.98$ \\
    Weight decay / gradient clipping & $0.1$ / $1.0$ \\
    Training precision & BF16 \\
    PPO lower / upper clipping & $0.20$ / $0.28$ (ratio interval $[0.80,1.28]$) \\
    Reference KL & K2 estimator, coefficient $10^{-3}$ \\
    Reward-side KL / entropy coefficient & $0$ / $0$ \\
    Loss aggregation & Global token mean \\
    \midrule
    Sampling temperature & $1.0$ \\
    Top-$p$ / top-$k$ & $1.0$ / disabled \\
    Maximum context length & 131,072 tokens; no additional fixed per-turn generation cap \\
    Maximum assistant turns & 100 \\
    Token interface & Token-in/token-out (TITO) \\
    TIS mode & Token-level weighting with out-of-range masking \\
    TIS acceptance interval & $[0.5,2.0]$; no batch weight normalization \\
    Dynamic sampling & Complete groups with reward standard deviation $>10^{-6}$ \\
    Infrastructure failure retries & At most two physical attempts per logical rollout \\
    \bottomrule
  \end{tabularx}
\end{table}

\parhead{Token alignment and importance sampling.}
TITO preserves sampled token IDs and token-level rollout log-probabilities
for trainable model outputs. The training backend consumes these token
sequences without retokenizing the complete trajectory. Prompts, tool
observations, inserted template tokens, and model-specific boundary
replacements are excluded from the training loss. Token-level truncated
importance sampling (TIS) weights the policy loss by the ratio between
training-backend old-policy and rollout-policy probabilities. We retain
weights in $[0.5,2.0]$ and mask tokens outside this interval rather than
clipping their weights to its endpoints. The TIS mask also applies to the
KL term, while the loss denominator remains the original valid-token count.

\parhead{Selective credit masking.}
We detect format and tool-interface errors,
repeated tool interactions without progress, and excessive within-turn
repetition of text or tool calls. For a flagged assistant turn, we set
positive advantages to zero on all model-generated tokens while retaining
negative advantages. Original rewards, subsequent recovery turns, and the
KL term remain unchanged.

\parhead{Dynamic sampling and failure handling.}
Dynamic sampling discards incomplete or constant-reward groups and samples
new groups to replenish the training batch. Physical retries address
recoverable infrastructure failures, not low-reward outcomes.

\subsection{Evaluation Details}
\label{app:evaluation-details}
\label{app:baseline-results}

The main comparison and RL study use BFCL V4, $\tau^2$-Bench, and Claw-Eval.
The data-construction ablations use Toolathlon-Verified, AutomationBench,
and Claw-Eval. Evaluation follows each benchmark's native harness and
official aggregate metric.

\parhead{External baseline results.}
The BFCL V4 and $\tau^2$-Bench scores of Qwen3.5-35B-A3B in
Table~\ref{tab:main_results} are taken from its official model card.
Its Claw-Eval average is reported in the Qwen3.5-35B-A3B comparison column
of the official Qwen3.6-35B-A3B model card. For $\tau^2$-Bench, Qwen follows
the official setup except in the airline domain, where it applies the fixes
specified in the Claude Opus 4.5 system card.

\FloatBarrier
\Needspace{8\baselineskip}
\section{Additional Experimental Results}
\label{app:additional-results}

\subsection{Credit-Assignment Granularity}
\label{app:credit-assignment}
With SFT initialization, task pool, reward verification, and positive-credit
masking fixed, we compare trajectory-level and atomic-turn GRPO.
The former assigns one advantage from the trajectory's mean checkpoint
reward to all turns; the latter computes advantages within atomic-turn
groups from binary task rewards. Atomic-turn GRPO improves every benchmark
at both scales (Table~\ref{tab:rl-gains}).

\Needspace{18\baselineskip}
\subsection{RL Training Curves}
\label{app:learning-dynamics}

\parhead{Reward reporting.}
For the 14B model, each training step samples 16 tasks. For reward reporting,
each task contributes 16 complete root-to-leaf trajectories reconstructed
from the final rollout layer, giving 256 trajectories per step.

\parhead{Trajectory aggregation.}
For the reward curve in Figure~\ref{fig:rl-learning-weft-14b}, we reconstruct
complete trajectories by backtracking from each leaf in the final rollout
layer to the root. Each trajectory's reward is the mean of its binary
checkpoint outcomes (Equation~\ref{eq:checkpoint-reward}). We average these
rewards within each task group and then across task groups in the batch.
This reported statistic reflects completion over the full task; training
advantages use the current atomic task's binary reward within each
atomic-turn group.

\section{MegaMCP Serving Evaluation}
\label{app:weft-megamcp}

\label{app:megamcp-benchmark}

We compare MegaMCP with local MCP serving inside E2B sandboxes using identical
task traces.  Local serving packages MCP source, initialized databases, and
server processes into each sandbox.  MegaMCP instead shares the servers while
isolating mutable state by session.  Both configurations use the same tool
interfaces, initial states, planned calls, and end-of-task cleanup.

\parhead{Workloads and timing boundaries.}
The cold-start comparison contains 100 tasks, 400 MCP sessions, and 3,791
planned tool calls, with sandbox creation and provisioning timed for both
configurations.  A larger workload contains 1,000 tasks, 4,000 sessions, and
39,864 planned calls, with up to 1,000 calls in flight.  In this larger
workload, local servers are prewarmed, while MegaMCP session provisioning is
included in the timed interval.  The cold-start comparison therefore measures
startup under matched timing boundaries; the larger workload measures
execution at rollout scale under the stated deployment conditions.

\begin{table}[!htb]
  \centering
  \small
  \setlength{\tabcolsep}{4.5pt}
  \renewcommand{\arraystretch}{1.08}
  \caption{\textbf{Data uploaded to agent sandboxes (MiB).}
  Workload data comprise the query, MCP source, and initialized databases;
  runner files contain the replay driver and dynamic configuration.}
  \label{tab:megamcp-upload}
  \resizebox{\textwidth}{!}{%
  \begin{tabular}{llrrrrrrr}
    \toprule
    Workload & Serving path & Query & MCP source & Initdb & Workload data &
      Runner files & Total & Workload reduction \\
    \midrule
    100 tasks, cold
      & Local & 0.315 & 45.908 & 25.195 & 71.418 & 2.865 & 74.282 & -- \\
    100 tasks, cold
      & MegaMCP & 0.315 & 0 & 0 & 0.315 & 2.893 & 3.208 & 99.559\% \\
    \addlinespace
    1,000 tasks, warmed
      & Local & 3.123 & 484.494 & 255.641 & 743.259 & 30.283 & 773.542 & -- \\
    1,000 tasks, warmed
      & MegaMCP & 3.123 & 0 & 0 & 3.123 & 31.717 & 34.841 & 99.580\% \\
    \bottomrule
  \end{tabular}%
  }
\end{table}

\parhead{Provisioning data.}
MegaMCP removes repeated delivery of MCP source and initialized databases
(Table~\ref{tab:megamcp-upload}).  At 1,000 tasks, workload upload falls
from 743.259 to 3.123 MiB, a 99.580\% reduction; the 100-task comparison
shows a similar 99.559\% reduction.  Including runner files, total upload
decreases by 95.496\% and 95.682\%, respectively.  Shared serving thus
reduces data transfer while preserving the same task queries and private
session state.

\begin{table}[!htb]
  \centering
  \small
  \setlength{\tabcolsep}{5pt}
  \renewcommand{\arraystretch}{1.08}
  \caption{\textbf{Serving latency.}
  TTFT is measured from the beginning of the timed interval to the first tool
  call; end-to-end latency additionally includes trace execution and cleanup.
  Latencies and wall time are in seconds.}
  \label{tab:megamcp-latency}
  \resizebox{0.94\textwidth}{!}{%
  \begin{tabular}{llrrrrr}
    \toprule
    Workload & Serving path & TTFT p50 & TTFT p95 & E2E p50 & E2E p95 &
      Wall time \\
    \midrule
    100 tasks, cold
      & Local & 10.455 & 14.460 & 24.271 & 79.673 & 332.469 \\
    100 tasks, cold
      & MegaMCP & 4.794 & 6.198 & 18.609 & 71.006 & 321.383 \\
    \addlinespace
    1,000 tasks, warmed
      & Local & 0.009 & 0.016 & 15.097 & 107.217 & 501.220 \\
    1,000 tasks, warmed
      & MegaMCP & 39.281 & 69.983 & 74.056 & 175.964 & 587.394 \\
    \bottomrule
  \end{tabular}%
  }
\end{table}

\parhead{Latency and lifecycle completion.}
Both configurations complete every task lifecycle.
Under matched cold-start timing,
MegaMCP reduces median time to the first tool call by 54.1\%, median
end-to-end latency by 23.3\%, and total wall time by 3.3\%
(Table~\ref{tab:megamcp-latency}).  At 1,000 tasks,
MegaMCP completes all lifecycles but incurs higher latency than the prewarmed
local servers, with its session provisioning still included in the measurement.
The results distinguish the reduction in repeated data transfer from the
remaining cost of provisioning concurrent sessions.

\Needspace{20\baselineskip}
\parhead{Historical cold and warmed runs.}
Table~\ref{tab:megamcp-history} reports historical cold and warmed runs
on the same set of 1,000 tasks. The cold-repeat bars in
Figure~\ref{fig:megamcp-summary}(c) use the two rows named \emph{Cold repeat},
whose recorded median TTFT values are 428.6\,s for local serving and
17.4\,s for MegaMCP. The complete primary comparisons and their timing
boundaries are reported in
Table~\ref{tab:megamcp-latency}.

\begin{table}[!htb]
  \centering
  \small
  \setlength{\tabcolsep}{3pt}
  \caption{\textbf{Historical 1,000-task serving runs.} Wall time and median
  TTFT are in seconds; stages retain the names in the experiment records.}
  \label{tab:megamcp-history}
  \begin{tabular*}{\linewidth}{@{\extracolsep{\fill}}llrr@{}}
    \toprule
    Stage & Serving & Wall (s) & TTFT p50 (s) \\
    \midrule
    Cold baseline & Local & 859.7 & 348.9 \\
    Early cold & MegaMCP & 278.5 & 22.7 \\
    Stabilized cold & MegaMCP & 548.1 & 22.2 \\
    Cold repeat & Local & 968.9 & 428.6 \\
    Cold repeat & MegaMCP & 550.3 & 17.4 \\
    \midrule
    First warmed & Local & 500.7 & 1.278 \\
    First warmed & MegaMCP & 625.9 & 39.7 \\
    Final warmed & Local & 501.2 & 0.009 \\
    Final warmed & MegaMCP & 587.4 & 39.3 \\
    \bottomrule
  \end{tabular*}
\end{table}

\Needspace{10\baselineskip}
\parhead{Resident memory.}
Figure~\ref{fig:megamcp-summary}(b) summarizes resident memory under process
sharing. For each configuration, we report memory use relative to a baseline
with a separate process for each session (Table~\ref{tab:megamcp-memory}).
The 50-session configuration occupies 110\,MiB in one process.

\begin{table}[H]
  \centering
  \small
  \setlength{\tabcolsep}{5pt}
  \caption{\textbf{MegaMCP resident memory.} Memory use is expressed as a
  percentage of the corresponding baseline, which runs a separate process
  for each session.}
  \label{tab:megamcp-memory}
  \begin{tabular*}{\linewidth}{@{\extracolsep{\fill}}lrrr@{}}
    \toprule
    Configuration & Processes & Relative memory (\%) & Reduction (\%) \\
    \midrule
    52 server instances & 8 & 22.4 & 77.6 \\
    50 sessions & 1 & 4.3 & 95.7 \\
    \bottomrule
  \end{tabular*}
\end{table}

\end{document}